\documentclass[10pt]{article}
\usepackage[table,dvipsnames]{xcolor}   
\usepackage{graphicx}
\usepackage{tikz}
\usepackage{pgfplots}
\pgfplotsset{compat=1.18}
\usepackage{booktabs}
\usepackage{multirow}
\usepackage{array}
\usepackage{arydshln}
\usepackage{colortbl}
\usepackage{calc}
\usepackage{multicol}
\usepackage{wrapfig}
\usepackage{algorithm2e}
\usepackage{url}
\usepackage[palatino,letter]{preprint}          
\usepackage[bottom]{footmisc}

\newcommand{\x}{\mathbf{x}}
\newcommand{\y}{\mathbf{y}}

\definecolor{darkgreen}{rgb}{0.0, 0.5, 0.5}

\title{GenoMAS: Guided Multi-Agent Programming for Scientific Discovery from Genomic Data}
\author{%
  Haoyang Liu \\
  University of Illinois \\
  Urbana-Champaign \\
  \texttt{hl57@illinois.edu}
  \And
  Yijiang Li \\
  University of California, \\
  San Diego \\
  \texttt{yijiangli@ucsd.edu}
  \And
  Haohan Wang \\
  University of Illinois \\
  Urbana-Champaign \\
  \texttt{haohanw@illinois.edu}
}
\hypersetup{pdftitle={GenoMAS: Guided Multi-Agent Programming for Scientific Discovery from Genomic Data},
  pdfauthor={Haoyang Liu, Yijiang Li, Haohan Wang},
  pdfkeywords={LLM agents, multi-agent systems, gene expression analysis, scientific discovery}}

\begin{document}
\maketitle

\begin{abstract}
Gene expression analysis holds the key to many biomedical discoveries, yet extracting insights from raw transcriptomic data remains challenging due to the complexity of multiple large, semi-structured files and the need for extensive domain expertise.
Current approaches either rely on inflexible workflows that fail in edge cases or fully autonomous agents that lack the precision required for rigorous scientific inquiry.
\textbf{GenoMAS} addresses this by presenting a team of LLM-based agents that integrates structured workflows with autonomous adaptability, orchestrating diverse specialized agents through typed message-passing protocols.
At its core lies a guided-planning framework where programming agents decompose task guidelines into Action Units and dynamically choose to advance, revise, bypass, or backtrack, maintaining logical coherence while adapting to the idiosyncrasies of genomic data.
On the GenoTEX benchmark, GenoMAS achieves 89.13\% CSC for data preprocessing and 60.48\% F$_1$ for identifying trait-associated genes, surpassing prior art by 10.61\% and 16.85\%.
Beyond transcriptomics, it outperforms strong agent baselines on ScienceAgentBench and DiscoveryBench, two independent bioinformatics benchmarks.
Qualitative analysis shows that GenoMAS identifies gene-phenotype associations that are biologically plausible and corroborated by the literature.
Our code is available at \url{https://github.com/Liu-Hy/GenoMAS}.\footnote{Project page: \url{https://liu-hy.github.io/GenoMAS/}}

\end{abstract}

\section{Introduction}
\label{sec: introduction}
Scientific research increasingly depends on complex computational analysis, yet the design and execution of such analysis remain labor-intensive, error-prone, and difficult to scale. From genomics~\citep{conesa2016survey, stark2019rna, wang2009rna}, materials discovery~\citep{dai2024AutonoMobile}, and drug development~\citep{swanson2024virtual, vichentijevikj2026prompt} to personalized medicine~\citep{ginsburg2018precision}, the analytical backbone of modern science involves highly structured, multi-step workflows written in code. These workflows must integrate raw or semi-structured data, apply statistical or mechanistic models, and yield interpretable results, all under the constraints of evolving domain knowledge, platform variation, and methodological rigor. Recent advances in large language models (LLMs) have led to rapid progress in general-purpose agents capable of decomposing tasks~\citep{wang2023survey, luo2025large, openai2023gpt4}, interacting with tools~\citep{qin2023toolllmfacilitatinglargelanguage, xu2023tool, ye2024tooleyes, huang2024metatoolbenchmarklargelanguage}, and producing executable code~\citep{dong2023selfcollaboration, he2025llmmas, xue2025improve}. These developments have raised the prospect that LLM-based agents might soon contribute meaningfully to scientific discovery by automating analysis pipelines, exploring hypotheses, or refining computational models~\citep{baek2024researchagent, li2024chain, schmidgall2025agent, hypothesis2025}. However, realizing this promise requires bridging a fundamental gap between general reasoning ability and the structured, precision-driven nature of scientific computation.

\begin{figure*}[ht]
    \centering
    \includegraphics[width=0.95\linewidth]{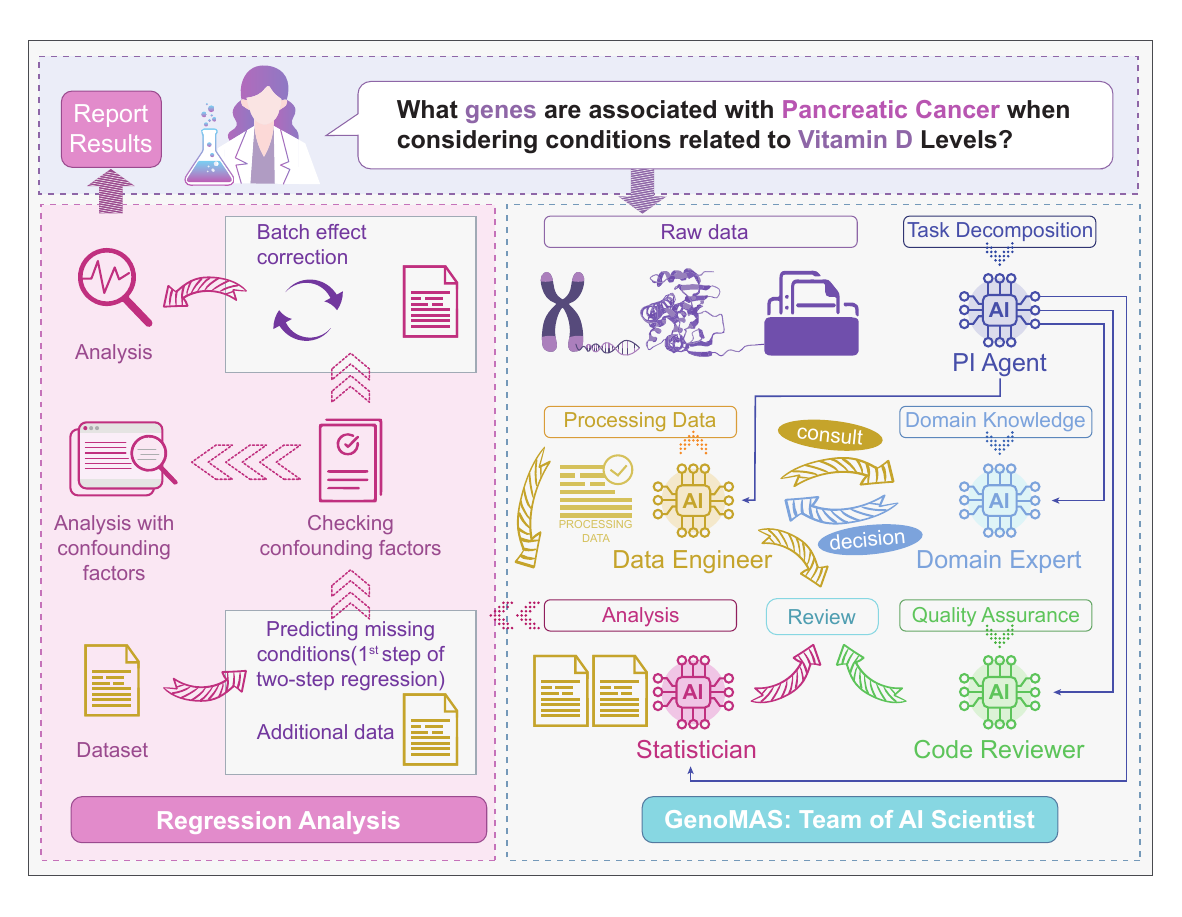}
    \caption{Multi-agent collaboration in our GenoMAS method.
    Given a user query, typically a request to identify disease-associated genes, the PI agent decomposes it into tasks;
    the Data Engineers preprocess the raw cohort data; and the Statistician performs the confounder-adjusted regression analysis.
    Throughout, the programming agents consult the Domain Expert for biology-dependent decisions and the Code Reviewer for code validation as needed, and the identified trait-associated genes are returned in an analysis report.
    }
    \label{fig:main}
\end{figure*}

While many LLM-based agents have shown competence in retrieving documents, calling APIs, or planning abstract tasks~\citep{zhuge2024gptswarm, zhang2024aflow, hu2024automated, wang2025scoreflow}, these capabilities fall short in domains where scientific progress depends on code. In fields such as transcriptomics~\citep{rosati2024differential, liu2024transcriptomics, ballard2024deep}, protein engineering~\citep{swanson2024virtual}, and statistical genetics~\citep{lander2025rare}, research workflows are encoded as sequences of programmatic transformations, each tailored to the idiosyncrasies of a specific dataset, model assumption, or experimental design. Analysts routinely write custom scripts to handle malformed metadata, adjust for confounding variables, map deprecated identifiers, and reimplement statistical procedures in response to batch effects or platform noise~\citep{johnson2007adjusting, khondoker2006statistical}. These operations cannot be abstracted into declarative graphs or offloaded to black-box tools; they require direct, editable code. Even in large biomedical consortia with standardized pipelines, deviations and adaptations remain common~\citep{conesa2016survey, stark2019rna, clough2016gene}. Scientific automation, therefore, hinges not merely on planning workflows, but on writing, revising, and validating the code that executes them.

Despite growing interest in agentic AI~\citep{wang2025llmagent, liu2025advances}, current systems rarely address this requirement. Most multi-agent frameworks operate by composing tool calls~\citep{qin2023toolllmfacilitatinglargelanguage, xu2023tool, huang2024metatoolbenchmarklargelanguage}, retrieving relevant APIs~\citep{qin2023toolllmfacilitatinglargelanguage}, or coordinating modular invocations across structured graphs~\citep{zhuge2024gptswarm, zhang2024aflow, hu2024automated, masgpt2025}. Some support dynamic routing~\citep{liu2023dynamic, saad2024archon}, temperature tuning~\citep{hu2024automated}, or neural module selection~\citep{zhang2025multi}, but even these operate within fixed architectural templates. A few systems attempt code-level control~\citep{Luyu2023PAL}, but lack structured revision mechanisms, validation layers, or domain-aware correction. Critically, tool-use benchmarks~\citep{yao2024taubenchbenchmarktoolagentuserinteraction, huang2024metatoolbenchmarklargelanguage} evaluate task completion or tool selection rather than scientific validity. In complex domains like gene expression analysis, even small errors in preprocessing or model selection can invalidate results entirely~\citep{leek2010tackling}. Yet current agent systems offer no reliable means to detect, debug, or recover from such errors once code is executed. As a result, they remain unsuitable for scientific workflows where correctness is inseparable from interpretability and domain context.

In this paper, to bridge the gap between general-purpose agentic reasoning and the domain-specific precision required for scientific computation, we present \textbf{GenoMAS} (\textbf{Geno}mic data analysis through LLM-based \textbf{M}ulti-\textbf{A}gent \textbf{S}ystem), a framework that reframes scientific agents as collaborative programmers rather than workflow orchestrators (Figure~\ref{fig:main}). Instead of invoking tools or chaining templates, GenoMAS agents generate, revise, and validate executable code tailored to each scientific task. Applied to gene expression analysis, the system automates complex workflows that previously required domain experts to script by hand. Evaluated on GenoTEX~\citep{liu2025genotex}, a benchmark of 1,384 gene-trait association tasks spanning heterogeneous cohorts, evolving metadata, and statistical confounders, GenoMAS achieves state-of-the-art performance while operating entirely from raw data inputs. The same design succeeds on ScienceAgentBench and DiscoveryBench, two independently developed benchmarks that span broader computational biology (Section~\ref{sec: generalization}). These results suggest that effective scientific automation requires agents that not only reason and retrieve, but that code.

Our contributions are as follows:
\begin{enumerate}
    \item We introduce GenoMAS, a multi-agent framework for scientific automation that treats agents as collaborative programmers rather than tool orchestrators, enabling end-to-end code generation for complex genomic analysis tasks.
    \item We develop a guided planning mechanism that encodes workflows as editable Action Units, structured textual directives that can be transformed into, and refined as, executable code, balancing precise control with autonomous error handling.
    \item We demonstrate that GenoMAS supports heterogeneous agent composition, allowing distinct LLMs with complementary strengths (e.g., code synthesis, language reasoning, scientific review) to operate in coordinated roles within the same execution loop.
\end{enumerate}

\section{Related Work}
\label{sec: related}
\paragraph{LLM-based Agents}
The emergence of LLMs \citep{openai2023gpt4, touvron2023llama, touvron2023llama2}, and of post-training methods that strengthen their reasoning \citep{reasoninglm2025, deepseek2025r1, li2026policy, wang2026learn}, 
has enabled the development of autonomous agents capable of complex reasoning and task execution \citep{wang2023survey, agentsurvey2025}. These agents leverage LLMs as their cognitive core, augmenting basic language capabilities with 
structured reasoning approaches and external tool use \citep{wang2022towards, wang2022self, hao2023reasoning, yao2022react}. Early agent methods explored decomposing  
complex tasks into manageable sub-goals \citep{wei2022chain, zheng2023synapse, feng2023towards, wang2022towards} and executing them sequentially. More
sophisticated agents organize reasoning into tree \cite{yao2023tree, hao2023reasoning} or graph structures \cite{besta2023graph}, enabling exploration of multiple solution 
paths. Critical to agent performance are mechanisms for self-reflection and iterative refinement \cite{xi2023self, madaan2023self, chen2023teaching, xue2025improve}, 
consistency checking \cite{wang2022self}, procedural knowledge packaged as reusable skills \cite{jiang2026demystifying}, and integration with external tools and knowledge bases \cite{liu2023llmp, zhao2023verifyandedit, qin2023toolllmfacilitatinglargelanguage, gou2023critic, 
qian2025toolrl}, which promises to transform LLMs from passive text generators into active problem-solving agents.  

\paragraph{Multi-Agent System} 
Given the capabilities of LLMs, multi-agent collaboration is expected to further enhance problem-solving performance 
\citep{wang2023unleashing, talebirad2023multiagent, du2023improving, gao2026localized, wang2023adapting}. In such systems, agents adopt specialized roles 
(i.e., role-playing) coordinated through structured protocols \citep{yang2023autogpt, dong2023selfcollaboration,yu2025sipdo}. For example, some agentic methods 
\citep{hong2023metagpt, qian2023communicative, dong2023selfcollaboration} organize agents into different roles emulating human collaboration for software development. 
Complementary strategies, such as goal decomposition and task planning \citep{wei2022chain, zheng2023synapse, feng2023towards, ning2024skeletonofthought,chen2025prompt}, and feedback mechanisms 
\citep{xu2023reasoning, gou2023critic, yin2023exchange, zhang2024revolve}, have also shown effectiveness. Beyond performance, recent work explores sociocognitive dynamics 
in multi-agent systems, revealing emergent social behaviors and theory-of-mind-like reasoning in simulated environments \citep{shapiro2023conceptual, sumers2023cognitive, 
zhou2023far, li2023theory, park2023generative}. More recent works have explored failure modes \cite{cemri2025why, zhang2025which} and training frameworks \cite{ma2024coevolving, masgpt2025, yang2026co} for multi-agent systems.

\paragraph{LLM Agents for Scientific Discovery} 
One of the most ambitious applications of LLM agents lies in scientific research, where they are being developed to assist, or even automate, various stages of the discovery
process \cite{lu2024ai, si2024can, schmidgall2025agent, baek2024researchagent, li2024chain, agentsurvey2025}, including peer review \cite{zhu2025deepreviewimprovingllmbasedpaper, 
weng2025cycleresearcher}. Recent systems demonstrate autonomous hypothesis generation \cite{hypothesis2025}, research assistance \cite{schmidgall2025agent}, and materials discovery \cite{llmatdesign2025}.
For instance, AI Scientist \cite{lu2024ai} demonstrates that frontier LLMs can independently complete an entire research cycle, 
while ResearchAgent \cite{baek2024researchagent} generates research problems, methodologies, and experimental designs, refining them iteratively using scientific literature. 
Recent efforts have also integrated LLMs into domain-specific inquiries in mathematics \cite{romera2024mathematical, thakur2023context, baba2025prover, deepseek2025r1}, 
physics \cite{ma2024llm, latif2024physicsassistant, zhang2025mooseagent}, chemistry \cite{sprueill2024chemreasoner, sprueill2023monte, bran2023chemcrow, guo2023can, llm34examples2025}, biology \cite{xiao2024proteingpt, 
madani2023large, zhou2023autoba, xiao2024cellagent, huang2025biomni}, and medicine \cite{singhal2023large, yang2023large, liu2024drugagent}, either through prompting or fine-tuning models on specialized datasets. 


\paragraph{Positioning of Our Work}
While existing agentic systems for scientific discovery have demonstrated capabilities in
hypothesis generation~\citep{baek2024researchagent}, experimental design~\citep{llmatdesign2025},
or literature-driven reasoning~\citep{kang2023comlittee, zhu2025deepreviewimprovingllmbasedpaper},
they rarely address settings where agents must write and revise executable code under scientific constraints.
GenoMAS targets this gap by treating scientific automation as a coding problem rather than a retrieval
or coordination problem.
Closest to our setting, a growing family of bioinformatics agents automates biological data analysis, including BioMaster~\citep{su2026biomaster}, CellAgent~\citep{xiao2024cellagent}, K-Dense Analyst~\citep{li2025kdense}, TransAgent~\citep{zhang2025transagent}, Biomni~\citep{huang2025biomni}, and GenoAgent~\citep{liu2025genotex}.
These systems share a planner-executor-reviewer pattern in which errors are repaired within the current step. GenoMAS builds on this pattern and differs in two mechanisms that Section~\ref{sec: experiment} evaluates. Guided planning over Action Units lets agents follow expert-written guidelines while choosing at each step whether to advance, revise, skip, or retract; retraction restores both the task context and the execution state to an earlier step once downstream evidence exposes a problematic decision, a global form of error recovery that current-step repair or replanning does not offer. Cross-family role specialization assigns complementary model families to the planning, programming, review, and domain reasoning roles, so that each agent contributes the particular strength of its backbone within one execution loop. Together, these mechanisms place GenoMAS between rigid workflows and fully autonomous agents, retaining the reliability of the former and the adaptability of the latter.

\section{Preliminaries}
\label{sec: preliminaries}
\subsection{Gene Expression Data Analysis}

Gene expression data analysis is a cornerstone of computational biology, offering critical insights into the molecular mechanisms that govern cellular function and disease \citep{rosati2024differential, byron2016rna, Love2014}. 
Advances in high-throughput technologies, such as microarrays \citep{abusamra2013comparative} and RNA sequencing \citep{wang2009rna, stark2019rna}, have enabled the generation of vast transcriptomic datasets, now housed in large-scale repositories including the Gene Expression Omnibus (GEO) \citep{clough2016gene, edgar2024ncbi} and The Cancer Genome Atlas (TCGA) \citep{tomczak2015cancer, weinstein2013cancer}.
This work centers on gene-trait association (GTA) analysis, a key task that identifies genes whose expression patterns 
correlate with specific phenotypic traits while rigorously controlling for confounding biological variables \citep{ghosh2005classification, wu2009genome}.

\subsection{Problem Formulation}

We address two classes of GTA problems following the GenoTEX benchmark \citep{liu2025genotex}:
\begin{itemize}
    \item \textbf{Unconditional GTA:} Given gene expression data $X \in \mathbb{R}^{n \times p}$ 
where $n$ is the number of samples and $p$ is the number of genes, and trait values $y \in \mathbb{R}^n$ (binary or continuous), 
identify the set of genes $G^* \subseteq \{1,...,p\}$ significantly associated with the trait.
    \item \textbf{Conditional GTA:} Given the above data and an additional condition variable $c \in \mathbb{R}^n$ (e.g., age, gender, or comorbidity), 
identify genes $G^*_c$ significantly associated with the trait when accounting for the influence of this condition \citep{kyalwazi2023race}. 
This conditional analysis helps explore direct gene-trait associations that exist independently of the condition's influence, 
enabling more precise insights for personalized medicine \citep{wang2022trade, chan2011personalized, hamburg2010path}.
\end{itemize}



The standardized analysis pipeline consists of three main stages: (1) dataset selection from available cohort studies, 
(2) data preprocessing to extract and normalize gene expression and clinical features, and 
(3) statistical analysis using regularized regression models to identify trait-associated genes \citep{cook2017guidance}.

\subsection{Evaluation}
\label{sec: evaluation}
We evaluate automated gene expression analysis methods using the GenoTEX benchmark \citep{liu2025genotex}, which provides standardized 
evaluation for each stage of the analysis pipeline described above. 
The framework assesses performance through three core tasks that mirror these stages:

\paragraph{Dataset Selection}
This task assesses a system’s ability to identify and prioritize relevant datasets for analysis.
Dataset Filtering (DF) measures binary classification accuracy in determining dataset relevance,
while Dataset Selection (DS) evaluates whether the method selects the same dataset as domain experts when multiple candidates are available.

\paragraph{Data Preprocessing}
Preprocessing entails extracting gene expression data from different platforms; microarray probes require
mapping to gene symbols \citep{khondoker2006statistical}, and RNA-seq reads demand alignment and quantification \citep{conesa2016survey, li2011rsem}.
Gene identifiers must be normalized across evolving nomenclatures and database versions \citep{sayers2022database, wei2023gnorm2, NCBIGene},
and clinical metadata extracted from heterogeneous semi-structured formats \citep{ccetin2022comprehensive}.
Preprocessing quality is evaluated using three metrics:
\begin{itemize}
    \item Attribute Jaccard (AJ) measures the overlap of extracted features (genes and clinical variables).
    \item Sample Jaccard (SJ) assesses the overlap of correctly integrated samples.
    \item Composite Similarity Correlation (CSC) = AJ $\times$ SJ $\times$ Corr$_{\text{avg}}$, where Corr$_{\text{avg}}$ 
is the average Pearson correlation coefficient between common features (shared rows and columns) of the preprocessed 
and reference datasets, capturing both structural and numerical fidelity.
\end{itemize}

\paragraph{Statistical Analysis}
Gene identification employs interpretable models, predominantly Lasso regression \citep{tibshirani1996regression} and its biologically informed variants \citep{wu2009genome, yang2024biologically}, which are well-suited for high-dimensional, sparse settings and yield compact, interpretable gene sets \citep{ghosh2005classification}.
Critical steps include batch effect correction (e.g., ComBat \citep{johnson2007adjusting} and its extensions \citep{zhang2020overcoming}), adjustment for population stratification \citep{yu2006unified, lippert2011fast}, and imputation of missing values \citep{liew2011missing, aittokallio2010dealing}.
Statistical analysis is evaluated primarily by the F$_1$ score of the identified gene set, complemented by Precision, Recall, AUROC \citep{hanley1982meaning}, which remains informative in this highly imbalanced setting (often fewer than 2\% of genes are trait-associated), and the Gene Set Enrichment Analysis (GSEA) Enrichment Score \citep{subramanian2005gene}.

\subsection{Technical Challenges}

Automated GTA analysis presents a set of deeply intertwined challenges that extend beyond conventional machine learning tasks \citep{angermueller2016deep}. 

\emph{(i) High-dimensional sparsity:} Gene expression datasets typically comprise over 20,000 genes measured across $<$1,000 samples, 
posing significant statistical hurdles \citep{johnstone2001distribution}, further exacerbated by biological noise and technical variability \citep{wang2020managing}. 

\emph{(ii) Platform heterogeneity:} Distinct measurement technologies demand distinct pipelines: microarrays rely on probe-based hybridization with platform-specific mappings \citep{abusamra2013comparative}, while RNA-seq requires complex alignment and quantification workflows \citep{stark2019rna, conesa2016survey}.

\emph{(iii) Evolving gene nomenclature:} The continuous evolution of gene names, marked by synonyms, deprecated identifiers, and context-specific aliases, necessitates robust normalization and disambiguation tools \citep{sayers2022database, wu2013biogps, wei2023gnorm2}.

\emph{(iv) Heterogeneous metadata:} Phenotypic information appears in varied formats, often requiring domain expertise 
to extract standardized variables from free-text descriptions or infer information from indirect sources \citep{chen2023seed}. 

\emph{(v) Confounding factors:} Batch effects \citep{leek2010tackling, wang2020managing}, population stratification \citep{yu2006unified}, 
and hidden covariates \citep{gagnon2012using, henderson1950estimation} can introduce spurious associations if not properly addressed \citep{bruning2016confounding}. 

Together, these challenges call for systems that combine advanced computational pipelines with nuanced biological understanding, enabling robust and generalizable GTA.

\section{Method: GenoMAS}
\label{sec: method}
\subsection{Overview}
The complexity of gene expression analysis demands both \emph{precise procedural control} and \emph{adaptive error handling}. It arises from the high-dimensional, noisy nature of genomic data and from the domain-specific decisions that critically affect downstream statistical outcomes.
Conventional workflow-based approaches offer structured execution but lack the adaptability to manage edge cases and persistent anomalies, leading to suboptimal performance and efficiency in our evaluations (Section~\ref{sec: results}); fully autonomous agents are more adaptable but sacrifice the precision essential for rigorous scientific analyses, and consistently fail to reach acceptable performance on benchmark tasks (Appendix~\ref{sec: failure_cases}).

GenoMAS reconciles these competing requirements through a multi-agent framework in which programming agents execute user-defined guidelines while retaining the autonomy to handle exceptions.
The system coordinates a team of specialized agents that communicate through structured protocols and employ context-aware decision-making to address edge cases, retry failed operations, and integrate domain-specific knowledge.

\subsection{Multi-Agent Architecture}

\paragraph{Agent Roles and Responsibilities}
GenoMAS employs six specialized agents organized into three functional categories.
(i) The \textbf{orchestration agent}, the \emph{PI agent}, coordinates the entire analysis workflow by dynamically assigning tasks based on analysis requirements and task dependencies.
(ii) The \textbf{programming agents} perform the core computational tasks in multi-step workflows: two \emph{Data Engineer agents} handle data preprocessing, with the \emph{GEO agent} focusing on GEO datasets and the \emph{TCGA agent} managing TCGA data, each specialized in its data formats and common preprocessing challenges; the \emph{Statistician agent} conducts the downstream statistical analyses, employing regression models to identify trait-associated genes while accounting for confounding factors.
(iii) The \textbf{advisory agents} support the programming agents through complementary expertise: the \emph{Code Reviewer agent} validates generated code for functionality and instruction conformance and suggests revisions, while the \emph{Domain Expert agent} provides biomedical insights for decisions requiring biological knowledge, such as clinical feature extraction and gene identifier mapping.

\paragraph{Diverse LLMs as Agents' Backbone}
Organizational science has established that cognitively diverse teams outperform homogeneous groups on complex tasks \citep{hong2004groups, page2007difference}. This insight extends to multi-agent systems, where combining models with complementary reasoning paths and roles yields stronger performance than any single model \citep{du2023improving, wang2024mixtureofagents, yang2026co}. Building on this principle, GenoMAS integrates a diverse set of state-of-the-art LLMs as backbones for its specialized agents. The programming agents are powered by Claude Sonnet 4 \citep{claude2024sonnet4}, selected for its strong agentic coding abilities. OpenAI o3 \citep{openai2025o3}, known for its robust reasoning capabilities, serves dual roles: guiding the planning logic of the programming agents and enabling the Code Reviewer to detect bugs and suggest targeted fixes. Gemini 2.5 Pro \citep{gemini2025pro}, one of the top-performing models on the GPQA \citep{rein2023gpqa} and HLE \citep{li2024hle} benchmarks, powers the Domain Expert agent with broad and accurate scientific knowledge, particularly in biology.

\paragraph{Task Orchestration}
The \textbf{PI agent} orchestrates analysis workflows by adhering to the dependency structure intrinsic to gene expression analysis. For each GTA task, it identifies cohorts that have already been preprocessed and schedules preprocessing only for the missing data, maximizing reuse across tasks with overlapping traits. Leveraging the independence of cohort-specific transformations, it parallelizes execution to reduce runtime. Upon completion of preprocessing, the \textbf{Statistician agent} selects the most suitable cohorts for downstream inference. Throughout the pipeline, the PI agent issues tasks via structured messages, monitors completion signals, and enforces configurable timeouts to prevent unbounded execution.

\paragraph{Message-Driven Task Progression}
GenoMAS employs a typed message-passing protocol that governs task progression through structured request-response cycles. Messages carry four key elements: sender role, message type (e.g., \texttt{CODE\_WRITING\_REQUEST}, \texttt{CODE\_REVIEW\_RESPONSE}), content, and target roles. For instance, upon completing a code segment, a programming agent sends a \texttt{CODE\_REVIEW\_REQUEST} targeting either the Code Reviewer or the Domain Expert, depending on the requirements of the current Action Unit. The receiving agent processes the request and returns a typed response that determines the next action: approval advances execution, while rejection triggers revision. A centralized environment manages these interactions via a message queue, enforcing topological constraints to avoid circular dependencies. This framework lets task flows emerge organically from agent interactions, while its structural rigor ensures coherent, goal-directed execution. Additional protocol details are provided in Appendix~\ref{sec: comm_protocol}.

\subsection{Programming Agents}
\begin{figure}[ht]
    \centering
    \includegraphics[width=0.8\linewidth]{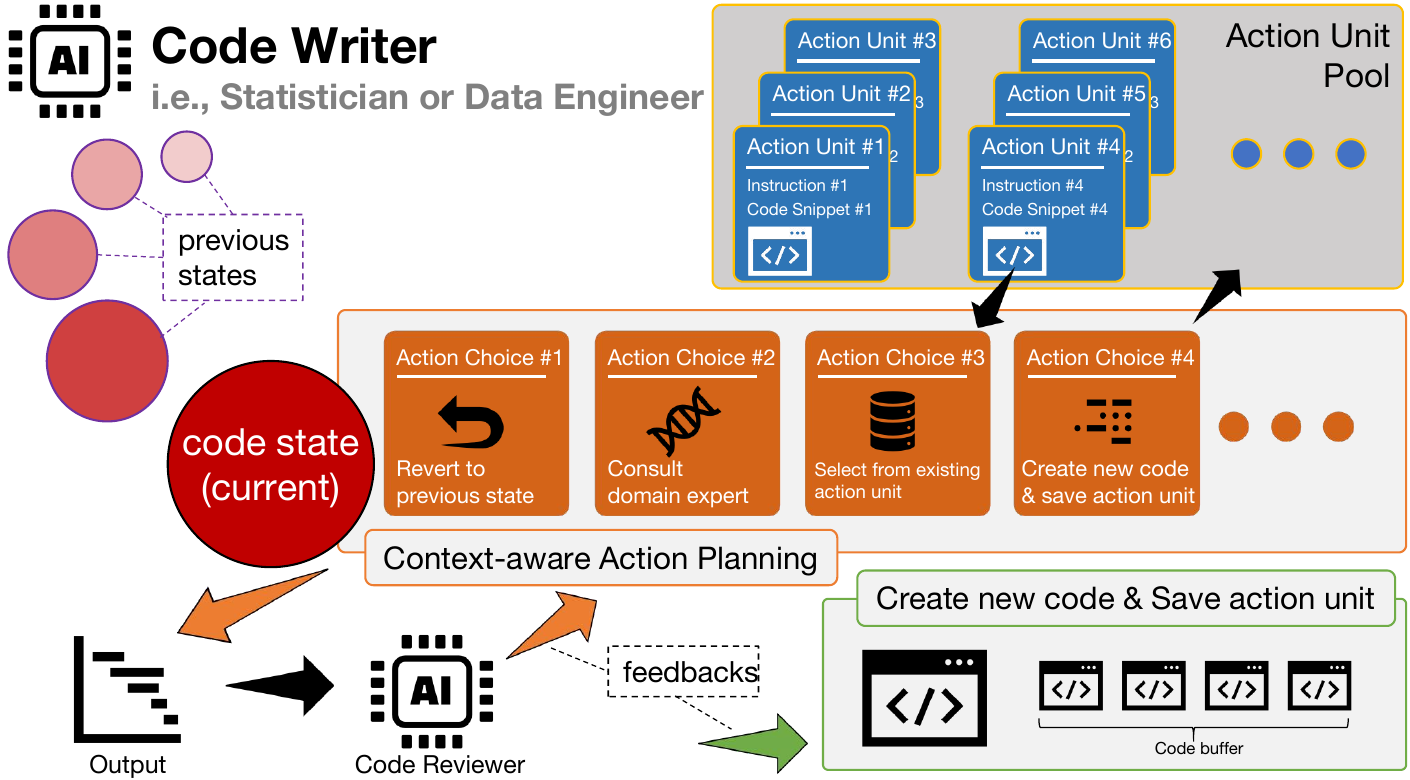}
    \caption{Planning, memory, and self-correction mechanisms of a single programming agent in our GenoMAS method.}
    \label{fig: programming_agent}
\end{figure}
\subsubsection{Guided Planning}
\paragraph{Guidelines and Action Unit}
The programming agents (\textbf{Data Engineers} and the \textbf{Statistician}) execute complex multi-step workflows using a guided planning framework that balances structure and flexibility. These agents operate under textual task guidelines that encode dependency structures, conditional logic, and termination criteria, effectively forming a directed acyclic graph (DAG). The guidelines serve as high-level execution blueprints, enabling agents to adjust their behavior dynamically based on the evolving context of each task.

Within this DAG, workflows are decomposed into \textit{Action Units}: semantically coherent operations that correspond to discrete subtasks. For example, the GEO agent's workflow comprises Action Units for data loading, clinical feature extraction, gene annotation, and normalization. Each unit represents a self-contained sequence of operations that can be executed atomically, without requiring intermediate oversight. The Action Units are initially generated by the agent from the guideline structure and subsequently refined through manual curation to ensure correctness and completeness. Appendix~\ref{sec: task_prompts} provides example guideline templates and Action Unit prompts for the different task categories.

\paragraph{Guided Context-aware Planning}
At each step, programming agents analyze their task history and current state to select the next Action Unit, considering the success or failure of previous steps, the data characteristics discovered during execution, and the remaining task objectives. Agents can advance to the next logical Action Unit, revisit a previous step with modified parameters, skip optional steps whose preconditions are not met, or terminate the workflow once its objectives are achieved.

The retraction mechanism allows agents to backtrack when they discover that an earlier decision led to downstream complications. Upon retraction, the agent reverts both its task context and its execution state to a previous step, then continues with an alternative Action Unit. This capability proves essential for handling the complex interdependencies in gene expression analysis, where early preprocessing decisions can have cascading effects on data quality. Appendix~\ref{sec: planning_mechanism} provides the metaprompts and further details of the planning framework.

\subsubsection{Domain Specific Code Generation}

\paragraph{Iterative Code Generation and Debugging}
GenoMAS adopts a three-stage process of code writing, review, and revision to generate robust analysis pipelines. For each Action Unit, programming agents receive the complete task context, including prior code executions, error traces, and historical attempts. This accumulated context allows them to reason about existing data structures and avoid redundant mistakes. The Code Reviewer evaluates the execution output, error messages, and adherence to the task guidelines, issuing either an approval or a detailed rejection. Based on this feedback, programming agents refine and resubmit their code, iterating until approval is granted or a predefined debugging limit is reached.

To ensure independent evaluation, the code review process enforces context isolation. The Code Reviewer receives the current code attempt and the overall task history, but not the feedback or decisions from previous review rounds at the same step. This design mitigates cascading biases and promotes objective assessments. After receiving a review response, the programming agent regains access to all prior attempts and feedback, allowing it to synthesize the insights and revise the code accordingly.

\paragraph{Domain Expert Consultation for Biomedical Reasoning}
For Action Units requiring biomedical knowledge, programming agents consult the Domain Expert agent in place of the Code Reviewer. The expert receives targeted context, such as metadata, processed summaries, and intermediate results, focusing on biological content rather than implementation details, and returns its guidance in executable code form, enabling contextually grounded, biologically valid operations. This process is also iterative: execution failures are routed back to the same expert for debugging, which keeps the reasoning consistent across refinement rounds. Details and example prompts are provided in Appendix~\ref{sec: code_generation}.

\paragraph{Code Memory for Reuse}
Programming agents maintain a dynamic memory of validated code snippets indexed by Action Unit type. Upon successful review, a snippet is stored for potential reuse in similar contexts. This memory evolves with experience, as agents can revise or replace stored code to reflect updated practices or domain shifts. The mechanism improves both efficiency and reliability by enabling the reuse of trusted patterns while preserving the flexibility to adapt to novel scenarios (Section~\ref{sec: memory}).

\paragraph{Specialized Tool Sets and Domain Knowledge Databases}
Each agent is equipped with a specialized set of Python tools aligned with its functional responsibilities. Data Engineers access utilities for dataset loading, DataFrame manipulation, and gene identifier mapping, while the Statistician agent employs statistical models and visualization libraries. These tools were co-developed by a doctoral student and a computational biologist through iterative testing on real genomic datasets, ensuring both robustness and domain relevance. Agents use them flexibly, invoking them directly or adapting them to new contexts within an Action Unit, which encapsulates best practices and simplifies code generation.

To ensure consistency and reduce external dependencies, GenoMAS integrates local, version-controlled biological knowledge bases. A curated gene synonym database from NCBI Gene \citep{NCBIGene} supports accurate symbol normalization across naming conventions, and gene-trait association data from the Open Targets Platform \citep{opentargets} enables the prediction of missing condition variables in conditional GTA tasks. The reference outputs used for evaluation are independent of Open Targets (see Section~\ref{sec: qualitative}). These resources are periodically updated to reflect current biomedical knowledge while ensuring reproducibility across experiments.

\subsection{System Implementation and Optimizations}
GenoMAS incorporates several system-level optimizations to support the practical demands of large-scale gene expression analysis, along three dimensions:
(i) \emph{Efficiency}: the system leverages asynchronous LLM calls to enable concurrent agent operations and adopts memory-efficient data processing strategies, such as streaming pipelines and selective column loading, to prevent out-of-memory failures on large genomic datasets;
(ii) \emph{Robustness}: a task management framework tracks completed analyses and supports automatic workflow resumption after interruptions, while real-time resource monitoring and configurable timeouts safeguard against runaway processes;
(iii) \emph{Scalability}: result caching and distributed task scheduling facilitate efficient execution across multiple GTA tasks.
Collectively, these design choices allow GenoMAS to process complex, real-world genomic data at scale while preserving the adaptability required for research-oriented exploration.

\section{Experiments}
\label{sec: experiment}
We evaluate \textbf{GenoMAS} on the GenoTEX benchmark \citep{liu2025genotex}, which encompasses 1,384 GTA problems across 913 datasets (41.5\,GB; 152,415 samples, 167 per cohort on average) spanning 132 human traits. GenoTEX covers the complete analytical workflow from raw data to biological insights, evaluates methods on real genomic datasets with realistic complexities rather than simplified or structured data, and provides ground truth curated and validated by professional bioinformaticians. These characteristics make it well suited for assessing agentic methods on complex, multi-step analysis that requires scientific rigor.

Conditional GTA predominates: 1,252 problems (90.5\%) are conditional and 132 (9.5\%) unconditional, with conditions comprising age (130), gender (121), and another biomedical trait (1,001).

\subsection{Experimental Setup}
\paragraph{Computation environment}
We conducted experiments on 6 RunPod GPU Cloud \citep{runpod} instances, each provisioned with 16 vCPU cores and 94GB RAM. The data analysis code generated by agents consumed approximately 0.3GB of CPU RAM during execution, consistently across LLM backbones. For LLM inference, proprietary models (e.g., OpenAI o3, Claude Sonnet 4) were accessed via their official APIs, while open-source models (DeepSeek R1, Qwen 3 235B) were served through Novita AI's infrastructure \citep{novitaai}, which offered lower latency than their official endpoints. Decoding settings and retry policies are documented in Appendix~\ref{sec: app_inference}.

\paragraph{Metrics} Our end-to-end evaluation assesses the ability of GenoMAS to identify trait-associated genes by analyzing raw input data in GTA tasks. We adopt the F$_1$ score (Section~\ref{sec: evaluation}) as the primary indicator of analytical performance, and additionally report Precision, Recall, AUROC, and the GSEA enrichment score. To validate the regression models produced by each method, we report trait prediction accuracy and F$_1$ for binary traits, obtained by using the fitted model to predict the trait on held-out samples of the datasets it selected.

In addition to task-specific metrics, we track several runtime indicators. The \emph{success rate} measures execution validity: the proportion of GTA problems for which the generated analysis code executes without errors and saves results to the correct path, irrespective of whether the identified genes are correct. \emph{Efficiency metrics} include input/output token counts, API cost, and average execution time per problem.

\subsection{Comparison with Prior Art}
We compare \textbf{GenoMAS} against the prior state-of-the-art method GenoAgent \citep{liu2025genotex}, against Biomni \citep{huang2025biomni}, a generalist biomedical agent, and against homogeneous variants of GenoMAS in which all agents share a single LLM backbone, testing several state-of-the-art models. GenoAgent and Biomni run on Claude Sonnet 4 (Thinking), the backbone of our programming agents, for a fair comparison. As empirical reference points, we report random gene selection and human expert performance from \citet{liu2025genotex}, the latter being the average performance of two independent expert analyses evaluated against a consensus ground truth. We also include a zero-shot baseline in which OpenAI o3 attempts to identify trait-associated genes without access to the data, probing whether parametric knowledge alone suffices for this task.

Table~\ref{tab: end2end_main} presents the end-to-end results. GenoMAS achieves a 60.48\% F$_1$ score, a 16.85\% absolute improvement over GenoAgent, alongside an AUROC of 0.81 (a 0.16 increase) and an approximately 45\% reduction in API cost. It attains 59.86\% F$_1$ on conditional and 66.37\% on unconditional GTA problems, and retains 84\% of the F$_1$ of the fully human-conducted analysis (71.63\%) without per-problem human intervention. The integration of heterogeneous LLMs proves essential: while Claude Sonnet 4 (Thinking) underpins our code generation, augmenting it with o3's reasoning abilities and Gemini 2.5 Pro's domain expertise yields a further 7.5\% F$_1$ improvement and an approximately 48\% cost reduction compared to the homogeneous Claude-only configuration. The o3 zero-shot baseline reaches only 5.13\% F$_1$ and 0.56 AUROC, confirming that the task demands structured data processing and domain reasoning rather than parametric knowledge alone.

Biomni, despite its success as a generalist biomedical agent built on comprehensive tool integration, reaches only 14.82\% F$_1$, even when given the same task guidelines and Action Unit prompts as GenoMAS (Appendix~\ref{sec: failure_cases}); workflow knowledge alone thus does not close the gap. The difference likely stems from design philosophy. Biomni relies on full agent autonomy to tackle diverse biomedical tasks with minimal user input, whereas we target the robust automation of more fully specified workflows with a higher demand for scientific rigor. Biomni therefore lacks the context management needed to plan each step from the task context and user guidelines while respecting the interdependencies between steps, and although its self-critic mechanism provides local quality control, it is not designed to detect and recover from systematic errors at the workflow level, where early methodological choices cascade through dozens of interdependent operations. Most other representative agents exhibited very low end-to-end success rates in preliminary experiments and are therefore excluded from Table~\ref{tab: end2end_main}, with the exception of Claude Code, which we quantify below.

\begin{table*}[t!]
\centering
\caption{\textbf{End-to-end performance of GenoMAS, prior art, a general-purpose coding agent, homogeneous-LLM variants, ablations, human experts, and data-free baselines.}
We report performance on GTA analysis, on the auxiliary trait prediction task, and on runtime metrics. We use $0$ as performance scores when code execution fails. ``Tk.(K)'', ``Cost (\$)'', and ``Time (s)''
represent average input/output tokens (in thousands), API cost, and runtime per problem, respectively. ``(T.)'' indicates
Extended Thinking mode.
}
\label{tab: end2end_main}
\renewcommand{\arraystretch}{1.18}
\fittable{
\footnotesize\setlength{\tabcolsep}{2.5pt}{
\begin{tabular}{p{3.4cm}ccccc@{\hspace{3mm}}cc@{\hspace{3mm}}cccc}
\toprule
 \multirow{2}{*}{Model} 
 & \multicolumn{5}{c@{\hspace{3mm}}}{GTA Analysis} 
 & \multicolumn{2}{c@{\hspace{3mm}}}{Trait Prediction} 
 & \multicolumn{4}{c}{Runtime Performance} \\
\cmidrule(l{2pt}r{2.5mm}){2-6} \cmidrule(l{2pt}r{3mm}){7-8} \cmidrule(l{2pt}r){9-12}
 & Prec.(\%) & Rec.(\%) & F$_1$(\%) & AUROC & GSEA & Acc.(\%) & F$_1$(\%) & Succ.(\%) & Tk.(K) & Cost(\$) & Time(s) \\
\midrule
\rowcolor[HTML]{E3F2FD}
GenoMAS (Ours)                                       & \textbf{61.75} & \textbf{66.67} & \textbf{60.48} & \textbf{0.81} & \textbf{0.95} & \textbf{86.25} & 71.34          & 98.78          & 162.6/7.1          & 0.38          & 307.26 \\
GenoAgent \cite{liu2025genotex}                      & 45.52          & 52.87          & 43.63          & 0.65          & 0.71          & 83.14          & 71.93          & 86.12          & 170.1/10.4         & 0.69          & 341.58 \\
Biomni \cite{huang2025biomni}                        & 17.42          & 15.24          & 14.82          & 0.29          & 0.39          & 48.54          & 41.64          & 53.66          & 213.4/9.3          & 0.78          & 183.91 \\
Claude Code (Sonnet 4 T.)                            & 9.22           & 8.61           & 8.53           & 0.10          & 0.11          & 9.40           & 7.36           & 12.14          & -                  & 0.70          & 340.24 \\
Claude Code (Sonnet 5 T.)                            & 48.92          & 51.82          & 47.34          & 0.69          & 0.83          & 75.32          & 62.89          & \textbf{99.42} & -                  & 0.35          & 483.29 \\
\addlinespace[1.7pt]
\hdashline[1.5pt/2pt]
\addlinespace[1.7pt]
Claude Sonnet 4 (T.) \cite{claude2024sonnet4}        & 53.36          & 55.88          & 52.98          & 0.74          & 0.93          & 85.77          & \textbf{73.60} & 96.34          & 177.4/13.2         & 0.73          & 282.85 \\
Claude Sonnet 4 \cite{claude2024sonnet4}             & 24.61          & 21.69          & 21.92          & 0.45          & 0.66          & 57.98          & 52.51          & 68.29          & 219.1/6.8          & 0.76          & 251.21 \\
OpenAI o3 \cite{openai2025o3}                        & 48.93          & 47.81          & 45.53          & 0.74          & \textbf{0.95} & 85.73          & 58.46          & 97.56          & 125.4/8.7          & 0.32          & 251.55 \\
OpenAI o3-mini \cite{openai2025o3mini}               & 39.08          & 37.46          & 35.54          & 0.52          & 0.57          & 64.07          & 38.77          & 74.39          & 136.7/10.6         & 0.20          & \textbf{127.37} \\
Gemini 2.5 Pro \cite{gemini2025pro}                  & 46.13          & 45.30          & 43.82          & 0.66          & 0.73          & 71.26          & 58.47          & 76.64          & 165.7/\textbf{3.3} & 0.24          & 293.96 \\
Gemini 2.5 Flash \cite{gemini2025flash}              & 40.29          & 43.26          & 40.67          & 0.56          & 0.59          & 50.14          & 33.82          & 61.70          & 142.0/\textbf{3.3} & 0.03          & 198.34 \\
DeepSeek R1 \cite{deepseek2025r1}                    & 13.42          & 13.58          & 13.45          & 0.13          & 0.14          & 11.23          & 8.95           & 13.94          & 107.9/28.7         & 0.15          & 851.23 \\
Qwen 3 235B \cite{qwen2025235b}                      & 6.73           & 6.89           & 6.52           & 0.07          & 0.09          & 8.14           & 5.47           & 9.28           & \textbf{78.0}/8.9  & \textbf{0.02} & 1084.51 \\
\addlinespace[1.7pt]
\hdashline[1.5pt/2pt]
\addlinespace[1.7pt]
No planning                                          & 51.61          & 57.01          & 51.27          & 0.74          & 0.87          & 73.32          & 59.90          & 89.59          & 204.1/9.3          & 0.48          & 359.79 \\
No Domain Expert                                     & 50.06          & 52.72          & 47.57          & 0.69          & 0.81          & 64.75          & 50.14          & 85.24          & 182.6/8.0          & 0.43          & 322.54 \\
Review Round=1                                       & 48.08          & 51.60          & 46.61          & 0.67          & 0.75          & 71.98          & 59.33          & 91.76          & 119.8/5.8          & 0.27          & 236.08 \\
No reviewer                                          & 25.12          & 26.89          & 24.98          & 0.45          & 0.52          & 47.43          & 30.84          & 56.23          & 96.6/3.6           & 0.19          & 138.36 \\
\addlinespace[2pt]
\hdashline
\addlinespace[2pt]
Human expert \cite{liu2025genotex}                   & 74.28          & 86.57          & 71.63          & 0.90          & 0.93          & -              &  -             &  -             &  -                 &  -            &  -     \\
\addlinespace[1.7pt]
\hdashline[1.5pt/2pt]
\addlinespace[1.7pt]
o3 zero-shot                                         & 15.84          &  4.28          &  5.13          & 0.56          & 0.41          & -              &  -             &  -              & 0.1/0.8           & 0.01          & 12.18 \\
Random \cite{liu2025genotex}                         &  0.72          &  0.90          &  0.75          & 0.49          & 0.04          & -              &  -             &  -              &  -                &  -            & 0.02 \\
\bottomrule
\end{tabular}
}
}
\end{table*}

\begin{table*}[ht!]
\centering
\caption{\textbf{Statistical analysis performance when using \emph{expert-preprocessed data} as input.} Metrics are as in Table~\ref{tab: end2end_main}. Baselines include representative agents and LLMs, and fixed scripts implementing two standard regression models. ``BEC'' in Row 2 represents batch effect correction.} 
\label{tab: regression}
\renewcommand{\arraystretch}{1.18}
\fittable{
\footnotesize\setlength{\tabcolsep}{2.5pt}{
\begin{tabular}{p{3.4cm}ccccc@{\hspace{3mm}}cc@{\hspace{3mm}}cccc}
\toprule
 \multirow{2}{*}{Model} 
 & \multicolumn{5}{c@{\hspace{3mm}}}{GTA Analysis} 
 & \multicolumn{2}{c@{\hspace{3mm}}}{Trait Prediction} 
 & \multicolumn{4}{c}{Runtime Performance} \\
\cmidrule(l{2pt}r{2.5mm}){2-6} \cmidrule(l{2pt}r{3mm}){7-8} \cmidrule(l{2pt}r){9-12}
 & Prec.(\%) & Rec.(\%) & F$_1$(\%) & AUROC & GSEA & Acc.(\%) & F$_1$(\%) & Succ.(\%) & Tk.(K) & Cost(\$) & Time(s) \\
\midrule
\rowcolor[HTML]{E3F2FD}
GenoMAS (Ours)                        & \textbf{97.14} & \textbf{92.87} & \textbf{95.26} & \textbf{0.97} & \textbf{0.99} & \textbf{88.37} & \textbf{77.21} & \textbf{100.00} & 25.0/1.7                  & 0.08          &  67.57 \\
GenoMAS (No BEC)                      & 70.18          & 75.60          & 69.64          & 0.85          & 0.92          & 86.48          & 61.29          & \textbf{100.00} & 24.2/1.6                  & 0.07          &  65.34 \\
GenoAgent \cite{liu2025genotex}       & 94.81          & 92.27          & 93.83          & 0.96          & 0.97          & 87.23          & 72.67          &  99.71          & 28.5/1.9                  & 0.09          &  71.60 \\
Biomni \cite{huang2025biomni}         & 90.56          & 88.56          & 89.23          & 0.92          & 0.94          & 85.05          & 71.87          &  96.52          & 26.1/1.4                  & 0.10          &  37.83 \\
Data Interpreter \cite{hong2024data}  & 91.20          & 89.74          & 90.86          & 0.94          & 0.97          & 88.10          & 73.54          &  99.16          & 43.0/2.8                  & 0.13          & 194.33 \\
MetaGPT \cite{hong2023metagpt}        & 87.28          & 88.73          & 87.46          & 0.91          & 0.94          & 84.88          & 70.35          &  96.05          & 36.2/2.4                  & 0.12          & 162.49 \\
AutoGen \cite{wu2023autogen}          & 86.04          & 87.09          & 86.38          & 0.93          & 0.95          & 82.81          & 72.24          &  96.60          & 41.7/2.6                  & 0.71          & 139.48 \\
CodeAct \cite{wang2024executable}     & 86.13          & 82.01          & 82.92          & 0.82          & 0.86          & 79.92          & 66.35          &  87.34          & 32.1/2.2                  & 0.11          & 110.93 \\
OpenAI o3 \cite{openai2025o3}         & 80.39          & 80.92          & 80.57          & 0.83          & 0.85          & 76.25          & 62.54          &  88.63          & 15.1/\textbf{1.5}         & 0.05          & \textbf{39.64} \\
DeepSeek R1 \cite{deepseek2025r1}     & 73.42          & 73.18          & 73.65          & 0.74          & 0.76          & 74.31          & 57.23          &  78.56          & \textbf{12.6}/4.9         & \textbf{0.02} & 136.22 \\
\addlinespace[1.7pt]
\hdashline[4.2pt/3pt]
\addlinespace[1.7pt]
Lasso \cite{tibshirani1996regression} & 31.88          & 12.99          & 14.03          & 0.53          & 0.12          & 76.48          & 62.31          &  99.39          & -                         & -             &  9.19 \\
LMM \cite{henderson1950estimation}    & 3.73           &  3.75          &  3.64          & 0.51          & 0.03          & 75.63          & 58.04          &  99.21          & -                         & -             &  15.44 \\
\bottomrule
\end{tabular}
}
}
\end{table*}

\paragraph{Comparison with a general-purpose coding agent}
\label{sec: general_agents}
To quantify the qualitative failures in Appendix~\ref{sec: failure_cases}, we ran Claude Code \citep{anthropic2024claudecode}, a widely used general-purpose coding agent, on all 1,384 problems under the protocol of Table~\ref{tab: end2end_main}, using one harness with two backbones: Claude Sonnet 4 (T.), matching GenoMAS's programming backbone and thinking budget, and the newer Claude Sonnet 5 \citep{claude2026sonnet5} with high-effort adaptive thinking (Appendix~\ref{sec: app_general_agent}). With the backbone fixed at Sonnet 4, Claude Code reaches 8.53\% F$_1$ at a 12.14\% success rate, versus 52.98\% and 96.34\% for homogeneous GenoMAS; as the two share backbone, guidelines, and metrics, this 44-point gap is attributable to the GenoMAS scaffold. Sonnet 5 lifts execution success to 99.42\%, the highest in the table, yet F$_1$ reaches only 47.34\%, below the GenoMAS scaffold on the older Sonnet 4 and far below full GenoMAS: backbone progress largely resolves execution completeness but does not yet close the gap in analysis accuracy that our framework contributes. These gaps confirm that reliable gene expression analysis requires combining precise procedural control with agentic autonomy; Appendix~\ref{sec: failure_cases} details the evaluation procedures and failure analyses.

\subsection{Ablation Study}
\label{sec: results}

\paragraph{Ablation on each component}
We conducted systematic ablations to quantify the contribution of each architectural component:
(i) removing the planning mechanism forces execution along a fixed workflow defined by the static DAG, where agents retain access to the task guidelines but cannot adapt the execution order;
(ii) excluding the Domain Expert agent tests whether the programming and Code Reviewer agents alone can handle biomedical reasoning;
(iii) restricting agents to a single review round assesses the impact of iterative refinement; and
(iv) omitting code review entirely measures performance without quality control.

The results in Table~\ref{tab: end2end_main} highlight the significance of each component. Context-aware planning enables dynamic adaptation to edge cases and error recovery, yielding both higher accuracy and greater efficiency by eliminating redundant steps and minimizing revision cycles. The collaborative design, particularly the specialized Code Reviewer and Domain Expert agents, proves critical to scientific rigor, and multiple review rounds catch subtle errors with downstream impact.

The ablations also separate the sources of accuracy from the sources of efficiency. Accuracy rests substantially on guided planning and on the Domain Expert: removing planning lowers F$_1$ by 9.2 points and removing the Domain Expert by 12.9 points, the latter reflecting the biomedical grounding the expert provides for clinical metadata extraction. Efficiency rests on planning as well as on code memory (Section~\ref{sec: memory}): without planning, agents fall into error loops, so the cost per problem rises from \$0.38 to \$0.48 and the runtime from 307 to 360 seconds despite the smaller system. Code review is the largest single factor for executable correctness. Without it, the success rate falls to 56.23\% and F$_1$ to 24.98\%, whereas a single review round already recovers most of the success rate (91.76\%) at a lower cost (\$0.27) but still leaves 13.9 F$_1$ points on the table.

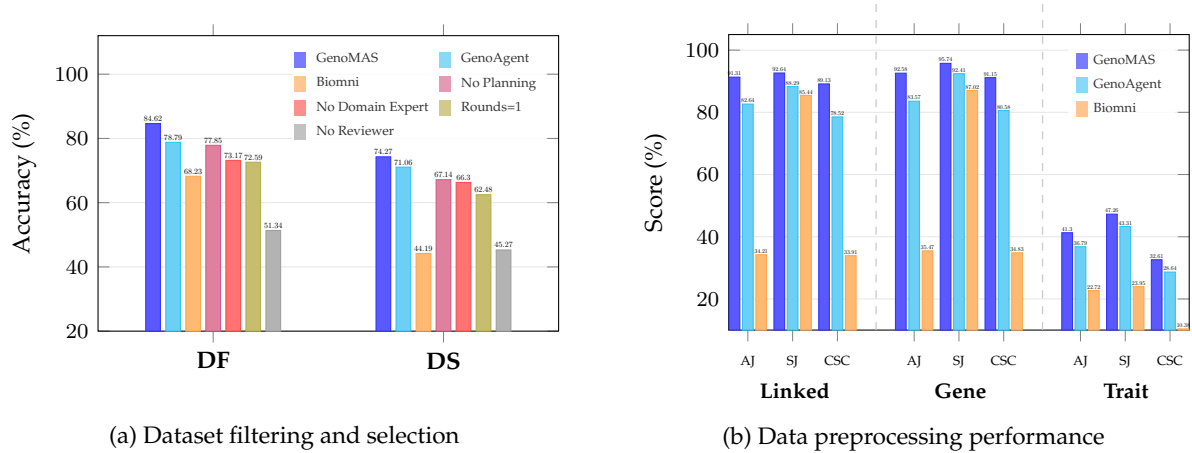
\begin{figure*}[t]
\centering
\begin{minipage}[t]{0.48\textwidth}
\centering
\begin{tikzpicture}[baseline=(current bounding box.north)]
\begin{axis}[
    ybar,
    bar width=5.5pt,
    width=\linewidth,
    height=5.5cm,
    ylabel={Accuracy (\%)},
    ylabel style={font=\small},
    symbolic x coords={DF,DS},
    xtick=data,
    xticklabel style={font=\small\bfseries},
    ymin=20,
    ymax=132,
    ytick={20,40,60,80,100},
    yticklabel style={font=\footnotesize},
    ymajorgrids=true,
    grid style={line width=.1pt, draw=gray!20},
    legend pos=north east,
    legend columns=2,
    legend image code/.code={
        \draw[#1,draw=none] (0cm,-0.07cm) rectangle (0.12cm,0.07cm);
    },
    legend style={
        font=\tiny,
        cells={anchor=west},
        draw=none,
        fill=white,
        fill opacity=0.8,
        column sep=0.2em,
        inner sep=1pt,
    },
    enlarge x limits=0.50,
    nodes near coords,
    nodes near coords align={vertical},
    every node near coord/.append style={inner sep=0pt, scale=0.30},
]

\addplot[
    fill=blue!65,
    draw=blue!80,
    every node near coord/.append style={yshift=3pt}
] coordinates {
    (DF,84.62)
    (DS,74.27)
};

\addplot[
    fill=cyan!50,
    draw=cyan!65,
    every node near coord/.append style={yshift=3pt}
] coordinates {
    (DF,78.79)
    (DS,71.06)
};

\addplot[
    fill=orange!55,
    draw=orange!70,
    every node near coord/.append style={yshift=3pt}
] coordinates {
    (DF,68.23)
    (DS,44.19)
};

\addplot[
    fill=purple!50,
    draw=purple!65,
    visualization depends on={y \as \YCoord},
    nodes near coords,
    every node near coord/.append style={
        yshift={ifthenelse(\YCoord>70,3pt,3pt)}
    }
] coordinates {
    (DF,77.85)
    (DS,67.14)
};

\addplot[
    fill=red!55,
    draw=red!70,
    visualization depends on={y \as \YCoord},
    nodes near coords,
    every node near coord/.append style={
        yshift={ifthenelse(\YCoord>70,3pt,3pt)}
    }
] coordinates {
    (DF,73.17)
    (DS,66.30)
};

\addplot[
    fill=olive!55,
    draw=olive!70,
    every node near coord/.append style={yshift=3pt}
] coordinates {
    (DF,72.59)
    (DS,62.48)
};

\addplot[
    fill=gray!60,
    draw=gray!75,
    every node near coord/.append style={yshift=3pt}
] coordinates {
    (DF,51.34)
    (DS,45.27)
};

\legend{GenoMAS, GenoAgent, Biomni, No Planning, No Domain Expert, Rounds=1, No Reviewer}

\node[font=\footnotesize\bfseries, anchor=north] at (axis description cs:0.500,-0.10) {\phantom{Gene}};
\end{axis}
\end{tikzpicture}
\subcaption{Dataset filtering and selection}
\end{minipage}
\hfill
\begin{minipage}[t]{0.48\textwidth}
\centering
\begin{tikzpicture}[baseline=(current bounding box.north)]
\begin{axis}[
    ybar=0.5pt,
    bar width=4pt,
    width=\linewidth,
    height=5.5cm,
    ylabel={Score (\%)},
    ylabel style={font=\small},
    xmin=-15,
    xmax=345,
    xtick={0,35,70,130,165,200,260,295,330},
    xticklabels={AJ,SJ,CSC,AJ,SJ,CSC,AJ,SJ,CSC},
    xticklabel style={font=\tiny},
    ymin=10,
    ymax=105,
    ytick={20,40,60,80,100},
    yticklabel style={font=\footnotesize},
    ymajorgrids=true,
    grid style={line width=.1pt, draw=gray!20},
    legend pos=north east,
    legend image code/.code={
        \draw[#1,draw=none] (0cm,-0.1cm) rectangle (0.15cm,0.1cm);
    },
    legend style={
        font=\tiny,
        cells={anchor=west},
        draw=none,
        fill=white,
        fill opacity=0.8,
    },
    x tick label style={yshift=-2pt},
    nodes near coords,
    nodes near coords align={vertical},
    every node near coord/.append style={inner sep=0pt, scale=0.22},
]

\addplot[
    fill=blue!65,
    draw=blue!80,
    bar shift=-5pt,
    every node near coord/.append style={yshift=3pt}
] coordinates {
    (0,91.31)
    (35,92.64)
    (70,89.13)
};

\addplot[
    fill=cyan!50,
    draw=cyan!65,
    bar shift=0pt,
    every node near coord/.append style={yshift=3pt}
] coordinates {
    (0,82.64)
    (35,88.29)
    (70,78.52)
};

\addplot[
    fill=orange!55,
    draw=orange!70,
    bar shift=5pt,
    every node near coord/.append style={yshift=3pt}
] coordinates {
    (0,34.21)
    (35,85.44)
    (70,33.91)
};

\addplot[
    fill=blue!65,
    draw=blue!80,
    bar shift=-5pt,
    forget plot,
    every node near coord/.append style={yshift=3pt}
] coordinates {
    (130,92.58)
    (165,95.74)
    (200,91.15)
};

\addplot[
    fill=cyan!50,
    draw=cyan!65,
    bar shift=0pt,
    forget plot,
    every node near coord/.append style={yshift=3pt}
] coordinates {
    (130,83.57)
    (165,92.41)
    (200,80.58)
};

\addplot[
    fill=orange!55,
    draw=orange!70,
    bar shift=5pt,
    forget plot,
    every node near coord/.append style={yshift=3pt}
] coordinates {
    (130,35.47)
    (165,87.02)
    (200,34.83)
};

\addplot[
    fill=blue!65,
    draw=blue!80,
    bar shift=-5pt,
    forget plot,
    every node near coord/.append style={yshift=3pt}
] coordinates {
    (260,41.30)
    (295,47.26)
    (330,32.61)
};

\addplot[
    fill=cyan!50,
    draw=cyan!65,
    bar shift=0pt,
    forget plot,
    every node near coord/.append style={yshift=3pt}
] coordinates {
    (260,36.79)
    (295,43.31)
    (330,28.64)
};

\addplot[
    fill=orange!55,
    draw=orange!70,
    bar shift=5pt,
    forget plot,
    every node near coord/.append style={yshift=3pt}
] coordinates {
    (260,22.72)
    (295,23.95)
    (330,10.38)
};

\legend{GenoMAS, GenoAgent, Biomni}

\draw[gray!40, dashed, line width=0.5pt] (axis cs:100,10) -- (axis cs:100,105);
\draw[gray!40, dashed, line width=0.5pt] (axis cs:230,10) -- (axis cs:230,105);

\node[font=\footnotesize\bfseries, anchor=north] at (axis description cs:0.139,-0.10) {Linked};
\node[font=\footnotesize\bfseries, anchor=north] at (axis description cs:0.500,-0.10) {Gene};
\node[font=\footnotesize\bfseries, anchor=north] at (axis description cs:0.861,-0.10) {Trait};
\end{axis}
\end{tikzpicture}
\subcaption{Data preprocessing performance}
\end{minipage}
\caption{\textbf{Individual task performance on dataset filtering, selection, and preprocessing.} 
(a) Dataset filtering (DF) and selection (DS) accuracy of different methods and GenoMAS ablation settings.
(b) Preprocessing performance across data types (Linked, Gene, Trait) and metrics (AJ: Attribute Jaccard; SJ: Sample Jaccard; CSC: Composite Similarity Correlation), with CSC primary.
}
\label{fig: combined_performance}
\end{figure*}

\paragraph{Individual Task Performance}
To identify performance bottlenecks, we evaluated the GenoMAS pipeline component-wise. For \emph{dataset filtering and selection} (Figure~\ref{fig: combined_performance}(a)), agents exhibit reasonable effectiveness, likely due to the comparatively lower reasoning complexity of metadata-based relevance assessment; however, early-stage errors can propagate through the pipeline and degrade overall performance.

\emph{Data preprocessing} (Figure~\ref{fig: combined_performance}(b)) reveals pronounced task-dependent variation, underscoring the inherent complexity of biological data analysis. GenoMAS achieves high performance on gene expression data (91.15\% CSC), demonstrating its effectiveness in handling genomic data transformation. In contrast, clinical trait preprocessing yields a markedly lower CSC of 32.61\%, reflecting the heterogeneous nature of clinical data and the nuanced domain knowledge required for accurate extraction. This disparity identifies clinical feature preprocessing as the main bottleneck, where both technical sophistication and biomedical expertise are essential. Even so, this CSC remains the highest among all evaluated systems; Section~\ref{sec: qualitative} and Appendix~\ref{sec: clinical_failures} examine the residual errors.

For \emph{statistical analysis} (Table~\ref{tab: regression}), we isolated this component by providing expert-preprocessed data as input, a controlled environment in which methods primarily need to apply standard statistical libraries. In this setting several agent-based approaches achieve competitive performance, yet important differences emerge. GenoMAS with batch effect correction achieves 95.26\% F$_1$, substantially outperforming traditional regression baselines (Lasso: 14.03\%), previous agent methods, and the variant without confound control. Basic statistical modeling is thus relatively straightforward for modern agents, but the methodological care to handle batch effects and covariate adjustment, which our system draws from its domain expertise, remains crucial for identifying genuinely significant biological signals.

\subsection{Generality Beyond Transcriptomics}
\label{sec: generalization}

While we focus on transcriptomic analysis, GenoMAS's mechanisms are generic by design. We therefore evaluate it on two independently developed bioinformatics benchmarks, holding the code-generating model fixed in the controlled comparisons so that differences reflect agent architecture. The full protocols are given in Appendix~\ref{sec: additional_experiments}.

\paragraph{ScienceAgentBench} Each of the 27 verified Bioinformatics tasks of ScienceAgentBench (SAB) \citep{chen2024scienceagentbench} requires writing and executing a program that produces a specified scientific artifact, judged by the task's own evaluator. GenoMAS (GPT-5 \citep{openai2025gpt5} for planning and review, Gemini 2.5 Pro as the Domain Expert, and Claude Sonnet 4.5 \citep{claude2025sonnet45} as the coder) is compared with Direct and Self-debug controls that share its coder, and with two public traces from the Holistic Agent Leaderboard (HAL) \citep{kapoor2026hal} re-scored under the same evaluator. As Table~\ref{tab: sab} shows, GenoMAS attains the highest success rate, 63.0\% (17/27), against 44.4\% for HAL SAB Self-Debug, 37.0\% for Direct, 29.6\% for Self-debug, and 25.9\% for the HAL generalist agent. Paired McNemar tests give significant gains over Direct ($p=.039$), Self-debug ($p=.004$), and the HAL generalist agent ($p=.002$); the 18.5-point gain over HAL SAB Self-Debug does not reach significance on 27 tasks ($p=.227$).

\begin{table}[t]
\centering
\caption{\textbf{Generality beyond transcriptomics.} (a) Success rates on the 27 verified Bioinformatics tasks of ScienceAgentBench, with 95\% Wilson intervals; CBS is SAB's secondary code-similarity metric. The two HAL rows are public traces re-scored under our evaluator, and Direct and Self-debug share GenoMAS's coder. (b) HMS on the 11 discovery-requiring DiscoveryBench biology tasks, averaged over three judge runs; all methods use Claude Sonnet 4.5 as the coding backbone and are scored by the same judge configuration.}
\label{tab: generality}
\captionsetup[subtable]{position=below, skip=4pt}
\footnotesize
\renewcommand{\arraystretch}{1.12}
\setlength{\tabcolsep}{5pt}
\begin{subtable}[t]{0.57\textwidth}
\centering
\begin{tabular}{lccc}
\toprule
       & Solved & SR (\%)   &     \\
Method & of 27  & [95\% CI] & CBS \\
\midrule
\rowcolor[HTML]{E3F2FD}
GenoMAS (ours)       & 17 & \textbf{63.0} [44.2, 78.5] & \textbf{0.870} \\
HAL SAB Self-Debug   & 12 & 44.4 [27.6, 62.7]          & 0.854 \\
Direct               & 10 & 37.0 [21.5, 55.8]          & 0.816 \\
Self-debug           &  8 & 29.6 [15.9, 48.5]          & 0.790 \\
HAL Generalist Agent &  7 & 25.9 [13.2, 44.7]          & 0.815 \\
\bottomrule
\end{tabular}
\caption{ScienceAgentBench}
\label{tab: sab}
\end{subtable}%
\hfill
\begin{subtable}[t]{0.40\textwidth}
\centering
\begin{tabular}{lc}
\toprule
Method & HMS \\
\midrule
\rowcolor[HTML]{E3F2FD}
GenoMAS (ours)             & \textbf{29.4} \\
\quad w/o guided planning  & 11.1 \\
\quad w/o code review      & 30.3 \\
Self-debug                 & 13.6 \\
DiscoveryBench coder agent &  8.1 \\
Direct (no analysis)       &  1.5 \\
\bottomrule
\end{tabular}
\caption{DiscoveryBench}
\label{tab: db}
\end{subtable}
\end{table}

\paragraph{DiscoveryBench} DiscoveryBench \citep{majumder2025discoverybench} evaluates a structurally different form of discovery: analyzing a dataset and stating the resulting hypothesis in natural language, scored against the gold hypothesis by an LLM judge (Hypothesis Matching Score, HMS, 0 to 100); its 20 public biology tasks concern freshwater-fish evolution and non-native plant introduction pathways. Because nine tasks embed the answer in the query (a no-analysis response already scores about 70 HMS), we report the 11 discovery-requiring tasks defined by Direct's failures (Table~\ref{tab: db}; Appendix~\ref{sec: app_db}). GenoMAS reaches 29.4 HMS, against 13.6 for Self-debug and 8.1 for DiscoveryBench's own coder agent on the same backbone. Removing guided planning drops GenoMAS to 11.1, into the range of the generic baselines, while removing code review is neutral (30.3): the margin here stems from guided planning, complementing the GenoTEX ablation, where review dominates because executable correctness is the bottleneck.

\subsection{Memory and Code Reuse Efficiency}
\label{sec: memory}

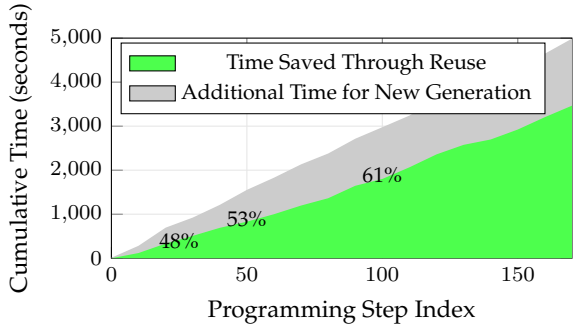
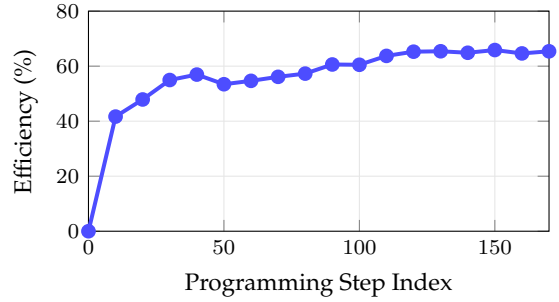
\begin{figure}[t!]
\centering
\begin{minipage}[t]{0.48\textwidth}
\centering
\begin{tikzpicture}
    \begin{axis}[
        width=\linewidth,
        height=4.5cm,
        xlabel={Programming Step Index},
        ylabel={Cumulative Time (seconds)},
        xlabel style={font=\small},
        ylabel style={font=\small},
        tick label style={font=\footnotesize},
        grid=major,
        grid style={gray!20},
        legend style={at={(0.02,0.98)}, anchor=north west, font=\scriptsize, inner sep=2pt},
        legend cell align=left,
        xmin=0, xmax=170,
        ymin=0, ymax=8000,
        xtick={0,50,100,150},
        ytick={0,2000,4000,6000,8000},
        yticklabel style={/pgf/number format/fixed}
    ]
    
    \addplot[fill=gray!40, draw=none, forget plot] coordinates {
        (0, 11.64)
        (10, 282.63)
        (20, 694.15)
        (30, 922.87)
        (40, 1212.47)
        (50, 1552.54)
        (60, 1829.14)
        (70, 2132.30)
        (80, 2378.58)
        (90, 2714.34)
        (100, 2975.86)
        (110, 3240.56)
        (120, 3614.36)
        (130, 3937.43)
        (140, 4086.12)
        (150, 4365.51)
        (160, 4646.44)
        (170, 4988.50)
    } \closedcycle;
    
    \addplot[fill=green!70, draw=none, forget plot] coordinates {
        (0, 0)
        (10, 117.78)
        (20, 332.40)
        (30, 507.20)
        (40, 690.42)
        (50, 829.57)
        (60, 999.80)
        (70, 1196.59)
        (80, 1362.72)
        (90, 1645.29)
        (100, 1800.45)
        (110, 2065.14)
        (120, 2358.80)
        (130, 2575.49)
        (140, 2697.22)
        (150, 2923.43)
        (160, 3204.36)
        (170, 3465.53)
    } \closedcycle;
    
    \addlegendimage{area legend, fill=green!70, draw=none}
    \addlegendentry{Time Saved Through Reuse}
    \addlegendimage{area legend, fill=gray!40, draw=none}
    \addlegendentry{Additional Time for New Generation}
    
    \node[font=\footnotesize] at (axis cs:25,400) {48\%};
    \node[font=\footnotesize] at (axis cs:50,900) {53\%};
    \node[font=\footnotesize] at (axis cs:100,1900) {61\%};
    
    \end{axis}
\end{tikzpicture}
\subcaption{Cumulative time savings}
\end{minipage}%
\hfill
\begin{minipage}[t]{0.48\textwidth}
\centering
\begin{tikzpicture}
    \begin{axis}[
        width=\linewidth,
        height=4.5cm,
        xlabel={Programming Step Index},
        ylabel={Efficiency (\%)},
        xlabel style={font=\small},
        ylabel style={font=\small},
        tick label style={font=\footnotesize},
        grid=major,
        grid style={gray!20},
        xmin=0, xmax=170,
        ymin=0, ymax=80,
        ytick={0,20,40,60,80},
        xtick={0,50,100,150}
    ]
    
    \addplot[color=blue!70, line width=1.5pt, mark=*, mark size=2pt, mark options={solid}] coordinates {
        (0, 0)
        (10, 41.67)
        (20, 47.89)
        (30, 54.96)
        (40, 56.94)
        (50, 53.43)
        (60, 54.66)
        (70, 56.12)
        (80, 57.29)
        (90, 60.61)
        (100, 60.50)
        (110, 63.73)
        (120, 65.26)
        (130, 65.41)
        (140, 64.87)
        (150, 65.86)
        (160, 64.63)
        (170, 65.40)
    };
    
    \end{axis}
\end{tikzpicture}
\subcaption{Memory reuse rate evolution}
\end{minipage}
\caption{\textbf{Memory reuse efficiency in GenoMAS.} (a) Cumulative time savings through memory reuse and (b) memory reuse rate evolution across programming steps. The system rapidly achieves high efficiency, stabilizing around 65\% reuse rate after initial learning.}
\label{fig: memory_efficiency}
\end{figure}

To understand how GenoMAS's memory mechanism contributes to its efficiency, we tracked code snippet reuse patterns during the analysis of the first 50 cohort datasets. This is a diagnostic analysis of a single tracked session, intended to illustrate how code reuse accumulates rather than to estimate benchmark-wide efficiency; the benchmark-wide cost and time figures are those in Table~\ref{tab: end2end_main}.
Figure~\ref{fig: memory_efficiency} demonstrates the efficiency gain: the system's dynamic memory of validated code snippets saves 57.8 minutes during the tracked session, with an average of 20.3 seconds per reused programming step. 
The memory reuse rate stabilizes around 65\% after initial learning, indicating that the system rapidly builds a reliable repertoire of reusable code patterns.

The memory mechanism's effectiveness arises because certain steps in gene expression analysis, such as loading GEO files, mapping gene symbols, and normalizing expression values, follow consistent patterns across datasets.
By capturing these patterns in reusable code snippets, GenoMAS transforms redundant code generation into efficient lookups, allowing the system to allocate computational resources to novel, cohort-specific challenges.

\section{Qualitative Analysis}
\label{sec: qualitative}
\paragraph{GenoMAS identifies biologically plausible gene-trait associations}
To assess whether GenoMAS produces scientifically interpretable results, we compared the genes it identified for the unconditional GTA problems against known gene-disease associations from the Open Targets Platform~\citep{opentargets}. The overlap yielded a GSEA enrichment score of 0.53 (FDR $q$-value 0.16), indicating moderate concordance with established disease biology. This comparison involves no circularity: GenoTEX's reference outputs are curated by human bioinformaticians independently of Open Targets, and GenoMAS itself consults Open Targets only to estimate missing condition variables in conditional problems, so no such queries arise for the unconditional problems examined here. Closer examination of high-confidence predictions for Alzheimer's disease, lung cancer, and hypertension reveals both well-established disease drivers (e.g., TERT for lung cancer) and biologically plausible candidates spanning disease-relevant mechanisms, from GABAergic synaptic dysfunction in Alzheimer's disease to Rho-kinase-mediated vascular tone regulation in hypertension. Exploratory per-gene literature evidence, documented in Appendix~\ref{sec: case_studies}, supports the use of GenoMAS outputs for hypothesis generation in genomic medicine.

\paragraph{Autonomous agent behaviors enhance workflow robustness}
Analysis of GenoMAS execution patterns reveals that agents autonomously adapt their behavior beyond prescribed instructions when encountering edge cases or persistent errors. Programming agents demonstrate context-aware problem-solving by proactively inserting diagnostic code (e.g., variable printing) to facilitate debugging, even without explicit reviewer guidance. In more complex cases, when an early decision creates downstream complications, agents can rewrite the code sequence from that decision point onward, correcting the execution trajectory without manual intervention. These emergent behaviors, documented in Appendix~\ref{sec: autonomy}, show how agent autonomy strengthens robustness against the inherent variability of genomic data.

To further leverage the autonomy of our agents, we prompt them to generate structured post-task notes categorized by severity (info, warning, error), documenting analysis challenges, data anomalies, and potential issues. This self-reporting mechanism enables human experts to audit system decisions efficiently by focusing on high-priority error notes, creating a feedback loop for continuous system improvement. Representative examples are provided in Appendix~\ref{sec: notes}.

\paragraph{Clinical feature extraction is the principal open challenge}
The same trajectories also make clear where the task remains hard. Turning free-text clinical metadata into a clean trait label is the most demanding part of the workflow, since it requires reconciling heterogeneous descriptions with exact phenotype definitions. GenoMAS attains the strongest clinical extraction performance among all evaluated systems (Section~\ref{sec: results}), yet a meaningful gap to expert curation remains. Reading through the trajectories, we group the residual errors into a few recurring judgment patterns: treating a within-patient contrast, such as controlled versus uncontrolled blood pressure in treated hypertensives, as a case-control label; drawing a trait from a cohort that does not study the target condition; entangling the trait with a covariate; and, rarely, synthesizing labels when extraction fails. The mechanical steps of preprocessing almost always run to completion; what fails is the judgment layered on top of them. Appendix~\ref{sec: clinical_failures} analyzes these cases with excerpts from the audited notebooks and outlines targeted safeguards, including a case-control validity check, a cohort relevance check, and a hard guard against synthetic labels, which we view as a promising direction for future work.

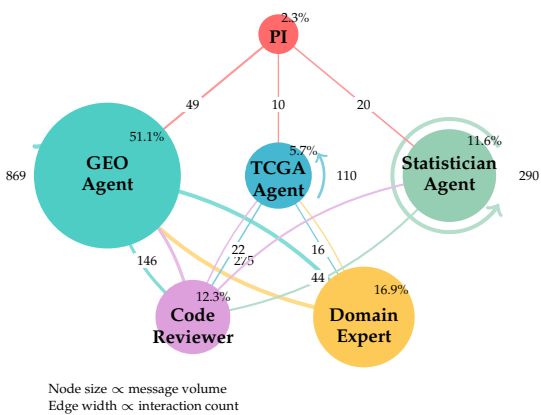
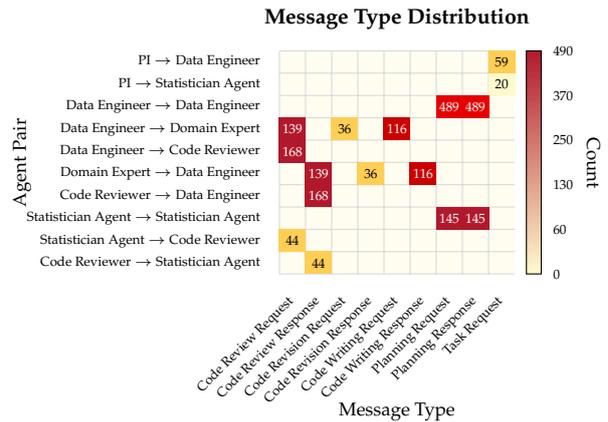
\begin{figure}[ht]
\centering
\begin{minipage}[t]{0.52\textwidth}
\raggedright
\begin{tikzpicture}[scale=0.75, every node/.style={font=\scriptsize}]
    \definecolor{picolor}{HTML}{FF6B6B}
    \definecolor{geocolor}{HTML}{4ECDC4}
    \definecolor{tcgacolor}{HTML}{45B7D1}
    \definecolor{statcolor}{HTML}{96CEB4}
    \definecolor{reviewcolor}{HTML}{DDA0DD}
    \definecolor{expertcolor}{HTML}{FECA57}
    
    \coordinate (PI) at (0,2.5);
    \coordinate (GEO) at (-3,0);
    \coordinate (TCGA) at (0,0);
    \coordinate (STAT) at (3,0);
    \coordinate (REV) at (-1.5,-2.5);
    \coordinate (EXP) at (1.5,-2.5);
    
    \draw[->, line width=2pt, geocolor!70] ([shift=(210:1.2)]GEO) arc (210:150:1.2);
    \node[font=\tiny, fill=white, inner sep=1pt] at ([shift=(180:1.6)]GEO) {869};
    
    \draw[->, line width=1.5pt, geocolor!70] (GEO) to[bend left=20] (EXP);
    \draw[->, line width=1.5pt, expertcolor!70] (EXP) to[bend left=20] (GEO);
    \node[font=\tiny, fill=white, inner sep=1pt] at (-0.6,-1.5) {275};
    
    \draw[->, line width=1.2pt, geocolor!70] (GEO) to[bend right=20] (REV);
    \draw[->, line width=1.2pt, reviewcolor!70] (REV) to[bend right=20] (GEO);
    \node[font=\tiny, fill=white, inner sep=1pt] at (-2.3,-1.5) {146};
    
    \draw[->, line width=1pt, tcgacolor!70] ([shift=(330:0.8)]TCGA) arc (330:390:0.8);
    \node[font=\tiny, fill=white, inner sep=1pt] at ([shift=(0:1.2)]TCGA) {110};
    
    \draw[->, line width=0.8pt, picolor!70] (PI) -- (GEO);
    \node[font=\tiny, fill=white, inner sep=1pt] at (-1.5,1.25) {49};
    
    \draw[->, line width=0.5pt, picolor!70] (PI) -- (TCGA);
    \node[font=\tiny, fill=white, inner sep=1pt] at (0,1.25) {10};
    
    \draw[->, line width=0.6pt, tcgacolor!70] (TCGA) -- (REV);
    \draw[->, line width=0.6pt, reviewcolor!70] (REV) to[bend left=10] (TCGA);
    \node[font=\tiny, fill=white, inner sep=1pt] at (-0.7,-1.3) {22};
    
    \draw[->, line width=0.5pt, tcgacolor!70] (TCGA) -- (EXP);
    \draw[->, line width=0.5pt, expertcolor!70] (EXP) to[bend right=10] (TCGA);
    \node[font=\tiny, fill=white, inner sep=1pt] at (0.7,-1.3) {16};
    
    \draw[->, line width=1.5pt, statcolor!70] ([shift=(30:1.0)]STAT) arc (30:330:1.0);
    \node[font=\tiny, fill=white, inner sep=1pt] at ([shift=(0:1.4)]STAT) {290};
    
    \draw[->, line width=0.6pt, picolor!70] (PI) -- (STAT);
    \node[font=\tiny, fill=white, inner sep=1pt] at (1.5,1.25) {20};
    
    \draw[->, line width=0.8pt, statcolor!70] (STAT) to[bend left=20] (REV);
    \draw[->, line width=0.8pt, reviewcolor!70] (REV) to[bend left=20] (STAT);
    \node[font=\tiny, fill=white, inner sep=1pt] at (0.7,-1.8) {44};
    
    \node[circle, fill=picolor, minimum size=15pt, inner sep=0pt] at (PI) {};
    \node[below=-5pt] at (PI) {\textbf{PI}};
    
    \node[circle, fill=geocolor, minimum size=55pt, inner sep=0pt] at (GEO) {};
    \node[text width=1.5cm, align=center] at (GEO) {\textbf{GEO}\\\textbf{Agent}};
    
    \node[circle, fill=tcgacolor, minimum size=25pt, inner sep=0pt] at (TCGA) {};
    \node[below=-9pt, text width=1.5cm, align=center] at (TCGA) {\textbf{TCGA}\\\textbf{Agent}};
    
    \node[circle, fill=statcolor, minimum size=35pt, inner sep=0pt] at (STAT) {};
    \node[text width=1.8cm, align=center] at (STAT) {\textbf{Statistician}\\\textbf{Agent}};
    
    \node[circle, fill=reviewcolor, minimum size=28pt, inner sep=0pt] at (REV) {};
    \node[below=-7pt, text width=1.5cm, align=center] at (REV) {\textbf{Code}\\\textbf{Reviewer}};
    
    \node[circle, fill=expertcolor, minimum size=38pt, inner sep=0pt] at (EXP) {};
    \node[below=-7pt, text width=1.5cm, align=center] at (EXP) {\textbf{Domain}\\\textbf{Expert}};
    
    \node[font=\tiny, inner sep=1pt] at ([shift=(45:28pt)]GEO) {51.1\%};
    \node[font=\tiny, inner sep=1pt] at ([shift=(45:18pt)]TCGA) {5.7\%};
    \node[font=\tiny, inner sep=1pt] at ([shift=(45:25pt)]STAT) {11.6\%};
    \node[font=\tiny, inner sep=1pt] at ([shift=(45:14.5pt)]REV) {12.3\%};
    \node[font=\tiny, inner sep=1pt] at ([shift=(45:19.5pt)]EXP) {16.9\%};
    \node[font=\tiny, inner sep=1pt] at ([shift=(45:12pt)]PI) {2.3\%};
    
    \node[font=\tiny, anchor=north west] at (-4.2,-3.5) {Node size $\propto$ message volume};
    \node[font=\tiny, anchor=north west] at (-4.2,-3.8) {Edge width $\propto$ interaction count};
\end{tikzpicture}
\subcaption{Agent collaboration network}
\end{minipage}%
\hspace{-20pt}
\begin{minipage}[t]{0.46\textwidth}
\raggedleft
\resizebox{\linewidth}{!}{\begin{tikzpicture}[scale=0.65, font=\small]
  \definecolor{low}{RGB}{255,250,205}     
  \definecolor{mid}{RGB}{255,206,84}      
  \definecolor{high}{RGB}{178,24,43}      
  \definecolor{background}{RGB}{255,253,240} 
  
  \foreach \y/\label in {
    9/{PI $\rightarrow$ Data Engineer},
    8/{PI $\rightarrow$ Statistician Agent},
    7/{Data Engineer $\rightarrow$ Data Engineer},
    6/{Data Engineer $\rightarrow$ Domain Expert},
    5/{Data Engineer $\rightarrow$ Code Reviewer},
    4/{Domain Expert $\rightarrow$ Data Engineer},
    3/{Code Reviewer $\rightarrow$ Data Engineer},
    2/{Statistician Agent $\rightarrow$ Statistician Agent},
    1/{Statistician Agent $\rightarrow$ Code Reviewer},
    0/{Code Reviewer $\rightarrow$ Statistician Agent}
  } {
    \node[anchor=east, font=\tiny] at (-0.2,\y*0.455+0.2275) {\label};
  }
  
  \foreach \x/\label in {
    0/{Code Review Request},
    1/{Code Review Response},
    2/{Code Revision Request},
    3/{Code Revision Response},
    4/{Code Writing Request},
    5/{Code Writing Response},
    6/{Planning Request},
    7/{Planning Response},
    8/{Task Request}
  } {
    \node[anchor=north east, rotate=45, font=\tiny] at (\x*0.529+0.15,-0.15) {\label};
  }
  
  \fill[background] (0,0) rectangle (4.76,4.55);
  
  \draw[gray!30] (0,0) grid[xstep=0.529,ystep=0.455] (4.76,4.55);
  
  \fill[mid] (8*0.529,9*0.455) rectangle (8*0.529+0.529,9*0.455+0.455);
  \node[font=\tiny] at (8*0.529+0.2645,9*0.455+0.2275) {59};
  
  \fill[low] (8*0.529,8*0.455) rectangle (8*0.529+0.529,8*0.455+0.455);
  \node[font=\tiny] at (8*0.529+0.2645,8*0.455+0.2275) {20};
  
  \fill[red!90!black] (6*0.529,7*0.455) rectangle (6*0.529+0.529,7*0.455+0.455);
  \node[white, font=\tiny] at (6*0.529+0.2645,7*0.455+0.2275) {489};
  \fill[red!90!black] (7*0.529,7*0.455) rectangle (7*0.529+0.529,7*0.455+0.455);
  \node[white, font=\tiny] at (7*0.529+0.2645,7*0.455+0.2275) {489};
  
  \fill[high] (0*0.529,6*0.455) rectangle (0*0.529+0.529,6*0.455+0.455);
  \node[white, font=\tiny] at (0*0.529+0.2645,6*0.455+0.2275) {139};
  \fill[mid] (2*0.529,6*0.455) rectangle (2*0.529+0.529,6*0.455+0.455);
  \node[font=\tiny] at (2*0.529+0.2645,6*0.455+0.2275) {36};
  \fill[red!80!black] (4*0.529,6*0.455) rectangle (4*0.529+0.529,6*0.455+0.455);
  \node[white, font=\tiny] at (4*0.529+0.2645,6*0.455+0.2275) {116};
  
  \fill[high] (0*0.529,5*0.455) rectangle (0*0.529+0.529,5*0.455+0.455);
  \node[white, font=\tiny] at (0*0.529+0.2645,5*0.455+0.2275) {168};
  
  \fill[high] (1*0.529,4*0.455) rectangle (1*0.529+0.529,4*0.455+0.455);
  \node[white, font=\tiny] at (1*0.529+0.2645,4*0.455+0.2275) {139};
  \fill[mid] (3*0.529,4*0.455) rectangle (3*0.529+0.529,4*0.455+0.455);
  \node[font=\tiny] at (3*0.529+0.2645,4*0.455+0.2275) {36};
  \fill[red!80!black] (5*0.529,4*0.455) rectangle (5*0.529+0.529,4*0.455+0.455);
  \node[white, font=\tiny] at (5*0.529+0.2645,4*0.455+0.2275) {116};
  
  \fill[high] (1*0.529,3*0.455) rectangle (1*0.529+0.529,3*0.455+0.455);
  \node[white, font=\tiny] at (1*0.529+0.2645,3*0.455+0.2275) {168};
  
  \fill[high] (6*0.529,2*0.455) rectangle (6*0.529+0.529,2*0.455+0.455);
  \node[white, font=\tiny] at (6*0.529+0.2645,2*0.455+0.2275) {145};
  \fill[high] (7*0.529,2*0.455) rectangle (7*0.529+0.529,2*0.455+0.455);
  \node[white, font=\tiny] at (7*0.529+0.2645,2*0.455+0.2275) {145};
  
  \fill[mid] (0*0.529,1*0.455) rectangle (0*0.529+0.529,1*0.455+0.455);
  \node[font=\tiny] at (0*0.529+0.2645,1*0.455+0.2275) {44};
  
  \fill[mid] (1*0.529,0*0.455) rectangle (1*0.529+0.529,0*0.455+0.455);
  \node[font=\tiny] at (1*0.529+0.2645,0*0.455+0.2275) {44};
  
  \node[font=\footnotesize\bfseries] at (2.38,5.2) {Message Type Distribution};
  
  \node[rotate=90, anchor=south, font=\scriptsize] at (-4.835,2.275) {Agent Pair};
  
  \node[font=\scriptsize] at (2.38,-2.8) {Message Type};
  
  \begin{scope}[shift={(5.0,0)}]
    \shade[top color=high, bottom color=low] (0,0) rectangle (0.3,4.55);
    \draw (0,0) rectangle (0.3,4.55);
    
    \node[anchor=west, font=\tiny] at (0.35,4.55) {490};
    \node[anchor=west, font=\tiny] at (0.35,3.64) {370};
    \node[anchor=west, font=\tiny] at (0.35,2.73) {250};
    \node[anchor=west, font=\tiny] at (0.35,1.82) {130};
    \node[anchor=west, font=\tiny] at (0.35,0.91) {60};
    \node[anchor=west, font=\tiny] at (0.35,0) {0};
    
    \node[rotate=-90, font=\scriptsize] at (1.359,2.275) {Count};
  \end{scope}
\end{tikzpicture}}
\subcaption{Message type distribution heatmap}
\end{minipage}
\caption{\textbf{Agent collaboration patterns in GenoMAS.} (a) Network topology showing agent communication structure with node size 
proportional to message volume. Edge thickness indicates interaction frequency. (b) Distribution of message types across agent pairs 
revealing asymmetric communication patterns, with programming agents predominantly sending validation requests while advisory agents 
respond with feedback. GEO and TCGA agents are merged into Data Engineer for brevity.}
\label{fig: collaboration_patterns}
\end{figure}

\paragraph{Collaboration patterns reflect role specialization and light-touch orchestration}
Figure~\ref{fig: collaboration_patterns} visualizes GenoMAS's agent communication structure during a representative 20-problem analysis session. The Data Engineer (the GEO and TCGA agents merged) accounts for 56.9\% of all interactions (1,956 messages), reflecting its central role in processing gene expression data, while the Statistician agent accounts for 11.6\% (398 messages) and the PI agent for only 2.3\%. Once tasks are dispatched, 97.7\% of the traffic is coordination among the programming and advisory agents rather than top-down control, with intensive bidirectional exchange between the Data Engineer and its two advisors. The heatmap shows that planning requests and responses (634 each) are the most frequent exchange, confirming that guided planning operates at every juncture of task execution; code validation requests (351) rank second, and revision requests are rare (36). We read these patterns descriptively. They show a design that concentrates execution in a few programming agents while distributing expertise across advisory agents with complementary backbones, in line with the cognitive-diversity rationale behind the heterogeneous architecture \citep{hong2004groups}, and in which the orchestrator rarely needs to intervene.

\section{Conclusion}
\label{sec: conclusion}
We presented \textbf{GenoMAS}, a multi-agent approach that advances automated gene expression analysis through three contributions:
guided planning that balances structured workflows with autonomous adaptation, heterogeneous LLM backbones whose complementary strengths in coding, reasoning, and domain knowledge are combined within one execution loop, and dynamic code memory that enables efficient reuse across analyses.
On the GenoTEX benchmark, GenoMAS reaches a 60.48\% F$_1$ score in gene-trait association analysis, a 16.85\% absolute gain over prior work, at approximately 45\% lower API cost, and the same design outperforms strong agent baselines on ScienceAgentBench and DiscoveryBench, two externally developed bioinformatics benchmarks. Its outputs include biologically plausible gene-phenotype associations that the literature corroborates.

The design principles behind GenoMAS, namely guided planning, cognitive diversity across backbones, and domain-informed programming, address a tension that is not specific to transcriptomics: scientific automation needs both precise procedural control and adaptive error handling. The generalization results suggest that these principles can carry over to other computational sciences in which correctness depends on executable code. At the same time, the remaining gap to expert analysts, concentrated in clinical feature extraction, marks the next problem to solve. Future work will explore multi-modal biological data integration, richer cross-task memory, more sophisticated planning algorithms, and the safeguards for clinical feature extraction outlined in Appendix~\ref{sec: clinical_failures}, while keeping the balance between automation and scientific trustworthiness that motivated this work.

\vspace{1.5\baselineskip}
\section*{Acknowledgements}
We thank Ye Zhang, Senior Expert Data Scientist at Novartis, for his valuable suggestions. This research was supported by the National AI Research Resource (NAIRR) under grant number 240283. This work used Microsoft Azure and Google Cloud through the CloudBank project~\citep{norman2021cloudbank}, which is supported by National Science Foundation grant \#1925001. This work also used CloudBank at UC San Diego through allocation CIS260773 from the Advanced Cyberinfrastructure Coordination Ecosystem: Services \& Support (ACCESS) program~\citep{boerner2023access}, which is supported by U.S. National Science Foundation grants \#2138259, \#2138286, \#2138307, \#2137603, and \#2138296.

\clearpage
\bibliographystyle{abbrvnat}
\bibliography{ref,ref2,ref3,ref4}

\appendix
\clearpage
\setcounter{section}{0} 
\onecolumn

\maketitlesupplementary

\renewcommand{\thesection}{\Alph{section}}

\definecolor{codebg}{RGB}{248, 249, 251}
\definecolor{retractiongray}{RGB}{120, 120, 125}
\lstset{style=preprint}

The supplementary material is organized as follows:

\begin{itemize}
\item Appendix \ref{sec: failure_cases} presents test cases showing fundamental limitations of existing autonomous agent methods on gene expression data analysis.
\item Appendix \ref{sec: additional_experiments} documents the protocols of the benchmark-wide evaluation of a general-purpose coding agent and of the two external benchmark evaluations, along with the inference configuration.
\item Appendix \ref{sec: comm_protocol} describes the communication protocols and messaging mechanisms in
GenoMAS.
\item Appendix \ref{sec: task_prompts} presents the task-specific guidelines and action unit prompts for GEO and TCGA preprocessing.
\item Appendix \ref{sec: planning_mechanism} details the guided planning framework and metaprompts for programming agents.
\item Appendix \ref{sec: code_generation} presents the code generation, review, and domain consultation mechanisms with their metaprompts.
\item Appendix \ref{sec: case_studies} presents exploratory biological plausibility case studies for selected GenoMAS gene-trait associations.
\item Appendix \ref{sec: autonomy} presents some examples of agent autonomy in GenoMAS.
\item Appendix \ref{sec: clinical_failures} analyzes recurring failure cases in clinical feature extraction and outlines targeted safeguards.
\item Appendix \ref{sec: notes} presents some examples of notes written by agents.
\end{itemize}

\section{Fundamental Limitations of Existing Autonomous Agent Methods}
\label{sec: failure_cases}

Recent advances in LLM-based agents have introduced numerous methods and products promising general-purpose automation with full autonomy. 
These systems handle task decomposition, planning, and multi-agent collaboration with minimal user input. To evaluate their applicability to gene expression analysis, 
we systematically tested six representative agents: three open-source systems (Biomni~\cite{huang2025biomni}, AutoGen~\cite{wu2023autogen}, 
MetaGPT~\cite{hong2023metagpt}) and three proprietary systems (Claude Code~\cite{anthropic2024claudecode}, Cursor~\cite{cursor2024}, Manus~\cite{shen2025manus}). 
For all agents, we use Claude Sonnet 4 (Thinking) as the underlying language model, and when configurable, set a thinking budget of 8192 tokens. This configuration represents 
the best performing setting in our end-to-end evaluation with homogeneous LLM configurations (achieving F$_1$ of 52.98\%), and most of these agents do not 
support heterogeneous multi-model configurations. Our experiments reveal that all these methods face fundamental limitations when applied to gene expression 
data analysis due to insufficient mechanisms for precise procedural control.

Among tested agents, Biomni incorporates 150 specialized biomedical tools, 105 software packages, and 59 databases with carefully designed tool retrieval mechanisms. 
This comprehensive biomedical specialization demonstrates the strongest domain adaptation among tested systems, yielding an F$_1$ score of 14.82\% in our 
experiments (Table~\ref{tab: end2end_main} in the main paper), substantially higher than other agents but still far below GenoMAS (60.48\%) and the previous best method GenoAgent (43.63\%); Section~\ref{sec: experiment} discusses the design differences behind this gap. The remaining five agents, 
designed for open-domain or general programming tasks, did not produce complete, valid analyses in our preliminary trials.
We subsequently evaluated one of them, Claude Code, end to end on all 1,384 benchmark problems (Section~\ref{sec: general_agents}; protocol in Appendix~\ref{sec: app_general_agent}). Its 12.14\% execution success rate on the Claude Sonnet 4 backbone confirms that the failure modes illustrated below are not confined to the selected cohorts.

Our analysis identifies two critical limitations in existing autonomous agents:

\textbf{Insufficient guidance interpretation}: General prompts prove inadequate for guiding agents through the complex analytical requirements of gene expression 
data. Agents frequently make consequential errors early in the workflow that propagate through subsequent steps, resulting in scientifically invalid outputs. 
This demonstrates the necessity of incorporating user-specified, domain-trusted guidelines into agent architectures.

\textbf{Absence of procedural control mechanisms}: Even when provided with detailed guidelines and Action Unit specifications, current agents lack mechanisms to 
ensure adherence to prescribed workflows. They cannot perform context-aware planning to identify deviations from expected procedures or recover from errors at the 
workflow level. This absence of precise procedural control renders them unsuitable for automation of gene expression analysis, where methodological rigor directly 
impacts scientific validity.

The remainder of this section presents examples and summary observations demonstrating these limitations in preprocessing GEO datasets. We focus on three cohorts for the trait
'Hypertension': GSE77627, GSE117261, and GSE256539. These cohorts contain complete information necessary for analysis and were successfully preprocessed by both human 
experts in the benchmark~\cite{liu2025genotex} and GenoMAS. For each agent, we provide results from two prompt conditions: (1) general instructions and (2) detailed 
instructions augmented with our guidelines and Action Unit specifications (provided in Appendix~\ref{sec: task_prompts}).
For readability, code blocks below are excerpts from the audited outputs and are not complete execution logs.

\subsection{Prompt Conditions}

\subsubsection{General Prompt}
We use the following general prompt, with minor modifications to adapt to the agent-specific I/O interface:
\begin{lstlisting}[language=TeX, escapeinside={(*@}{@*)}]
Please help me preprocess the cohort dataset.
I need to get a complete, linked dataset where each row represents a sample and each column represents a feature.
The row index should be the sample ID. The columns should include and only include the following:
- The trait represented as a binary or continuous variable, whichever is more suitable. If it is binary, use 1 to represent positive and 0 as control. Use the trait name as the column name.
- The sample's age (if available), a float or integer representing the age in years. Use "Age" as the column name.
- The sample's gender (if available), a binary variable with 0 for female and 1 for male. Use "Gender" as the column name.
- The expression values of genes, with gene symbols as column names.
\end{lstlisting}

\subsubsection{Detailed Prompt}
The general prompt augmented with our task-specific guidelines, followed by the reminder:
\begin{lstlisting}[language=TeX, escapeinside={(*@}{@*)}]
Below are more detailed descriptions about some of the steps in the guidelines. You are free to selectively execute all or some of them based on task context, and decide the order that is best for completing the task in accordance with the above guidelines.
\end{lstlisting}
This reminder is then followed by the Action Unit instructions, emphasizing agents' flexibility to plan and selectively execute based on context.

\subsection{Biomni Examples}

Despite incorporating specialized biomedical tools, Biomni demonstrates fundamental limitations in understanding disease biology 
and correctly interpreting clinical phenotypes. The following example from GSE77627 illustrates how these limitations manifest 
in both prompt conditions.

\subsubsection{GSE77627: Misinterpretation of Portal Hypertension Phenotypes}

This dataset studies portal hypertension, a condition characterized by elevated blood pressure in the portal venous system. 
The study includes three liver groups: HNL (Histologically Normal Livers), INCPH (Idiopathic Non-Cirrhotic Portal Hypertension), 
and LC (Liver Cirrhosis). Both INCPH and LC represent different etiologies of portal hypertension, while HNL serves as the 
healthy control group.

Under the general prompt condition, Biomni implements the following erroneous trait conversion:

\begin{lstlisting}[language=Python, escapeinside={(*@}{@*)}]
def convert_trait(value):
    """Convert trait values to binary (INCPH=1, others=0)"""
    if value is None:
        return None
    # Extract value after colon
    if ':' in str(value):
        val = str(value).split(':')[1].strip()
    else:
        val = str(value).strip()
    
    # Convert INCPH to 1 (positive), others to 0 (control)
    if val == 'INCPH':
        return 1
    elif val in ['HNL', 'LC']:  # Critical error: LC is portal hypertension, not control (*@\textcolor{red}{<--}@*)
        return 0
    else:
        return None
\end{lstlisting}

This trait assignment is misaligned with the benchmark phenotype definition used for this cohort. Liver cirrhosis accounts for
approximately 90\% of portal hypertension cases in Western countries and represents the most common etiology of this 
condition. For a portal-hypertension trait label, assigning LC samples to the control group substantially alters the
statistical composition of the dataset. The resulting classification creates a dataset with only
18 cases (INCPH alone) versus 36 controls (HNL and LC combined), whereas the benchmark phenotype definition yields
40 cases (both INCPH and LC) versus 14 controls (HNL only). This reversal in the case-control ratio
would substantially change the downstream analysis. Rather than identifying genes associated
with portal hypertension as defined in the benchmark, the analysis would instead detect genes that distinguish INCPH from a heterogeneous
mixture of healthy and cirrhotic livers, making the result difficult to interpret for the intended task.

The error persists even under the detailed prompt condition, where explicit guidelines and domain knowledge are provided. 
In the general prompt condition, Biomni compounds the misclassification by entirely excluding LC samples from the analysis:

\begin{lstlisting}[language=Python, escapeinside={(*@}{@*)}]
# From Biomni's general prompt processing
trait_mapping = {
    'HNL': 0,    # Control - histologically normal livers
    'INCPH': 1,  # Disease - idiopathic non-cirrhotic portal hypertension
    'LC': 'exclude'  # Exclude liver cirrhosis samples (*@\textcolor{red}{<--}@*)
}

# Filter samples and create sample metadata
final_sample_metadata = {}
for sample_id, metadata in sample_metadata.items():
    liver_group = metadata['liver_group']
    if liver_group != 'LC':  # Exclude LC samples (*@\textcolor{red}{<--}@*)
        final_sample_metadata[sample_id] = {
            'INCPH': trait_mapping[liver_group],
            'liver_group': liver_group
        }

print(f"Samples after filtering (excluding LC): {len(final_sample_metadata)}")
# Output: Samples after filtering (excluding LC): 32
\end{lstlisting}

This exclusion strategy eliminates 22 LC samples from the original 54, representing a 40\% data loss that substantially
reduces statistical power. The decision to exclude rather than include LC samples as portal-hypertension cases suggests a
failure to align the preprocessing logic with the benchmark phenotype definition.

Beyond the trait misclassification, Biomni exhibits several additional processing errors that compound the analysis 
invalidity. The agent fails to normalize gene symbols using standardized databases, a critical step performed by human 
experts to ensure cross-dataset compatibility. While the human expert code systematically applies gene synonym mapping to 
standardize identifiers, Biomni retains the raw gene symbols from the platform annotation, potentially introducing 
inconsistencies in downstream analyses. Furthermore, Biomni omits systematic bias detection and correction procedures. The 
human expert implementation includes explicit checks for biased features and removes them before final analysis, whereas 
Biomni proceeds without such quality control measures.

These cascading errors, from phenotype interpretation errors to technical processing omissions, illustrate that even
specialized biomedical tools cannot compensate for the lack of precise procedural control required for valid scientific 
analysis. The inability to recognize that different diseases can manifest the same clinical phenotype, combined with the 
failure to implement standard bioinformatics quality control procedures, renders the autonomous agent's output scientifically 
unreliable.

\subsection{AutoGen Summary}

AutoGen, a popular multi-agent conversation framework developed by Microsoft~\cite{wu2023autogen}, 
demonstrates fundamental limitations when applied to gene expression analysis. While designed to 
facilitate autonomous agent collaboration through conversational patterns, AutoGen struggles with the 
precise procedural control required for bioinformatics workflows. In our preliminary trials, AutoGen did not produce a
complete, valid preprocessing output for the tested GEO cohorts under either the general prompt or the detailed prompt.
The failures were consistent with the broader limitation observed across general-purpose agents: the framework could
coordinate subtasks, but it did not reliably enforce the domain-specific checks needed for phenotype interpretation,
probe-to-gene mapping, and final output validation.

Because the AutoGen runs did not yield concise, auditable execution traces suitable for reproduction in the appendix, we
do not include reconstructed code snippets or conversational examples here. We therefore use the AutoGen results only as
qualitative evidence that general multi-agent orchestration alone is insufficient for this setting, rather than as a
separate quantitative baseline for end-to-end raw-data preprocessing.

\subsection{MetaGPT Summary}

MetaGPT, a multi-agent framework emphasizing structured role-based collaboration~\cite{hong2023metagpt}, 
exhibits similar fundamental limitations despite its sophisticated role definitions and standardized 
operating procedures. The framework's emphasis on software engineering patterns fails to translate 
effectively to scientific data analysis requirements. In our preliminary trials, MetaGPT did not complete the standardized
GEO preprocessing workflow with outputs that satisfied the benchmark requirements. The observed failures reflected the
same pattern seen in other general-purpose systems: role-based decomposition and generated documentation did not ensure
that all required biological and technical checks were performed.

As with AutoGen, we omit detailed MetaGPT code examples because the available run artifacts are not suitable for concise
verbatim presentation in this appendix. We retain the qualitative observation because it motivates GenoMAS's emphasis on
explicit action units, code review, and domain consultation, but we do not treat MetaGPT as a quantitative baseline for
end-to-end raw-data preprocessing.

\subsection{Claude Code Examples}

Claude Code, arguably today's best general-purpose CLI-based agent, demonstrates similar fundamental limitations.  
This proprietary agent excels at general software development tasks but struggles with domain-specific scientific data analysis. 
The following examples illustrate how lack of domain knowledge and procedural control leads to critical errors in both 
prompt conditions.

\subsubsection{GSE77627: Misclassification of Portal Hypertension Subtypes}

Under the general prompt condition, Claude Code implements a three-class classification scheme that
is inconsistent with the benchmark's binary phenotype definition for portal hypertension:

\begin{lstlisting}[language=Python, escapeinside={(*@}{@*)}]
# Claude Code's trait conversion logic
def preprocess_GSE77627():
    # ... earlier code omitted ...
    
    # Convert liver groups to binary/categorical encoding
    # HNL (healthy normal liver) = 0 (control)
    # INCPH (idiopathic non-cirrhotic portal hypertension) = 1 (case)
    # LC (liver cirrhosis) = 2 (another condition)  (*@\textcolor{red}{<--}@*)
    group_mapping = {'HNL': 0, 'INCPH': 1, 'LC': 2}
    metadata_df['trait'] = metadata_df['liver_group'].map(group_mapping)
    
    print("Sample distribution:")
    print(metadata_df['liver_group'].value_counts())
    
    # Note: Age and Gender are not available in this dataset
    print(f"\nFinal dataset shape: {final_df.shape}")
    print(f"Columns: trait + {final_df.shape[1]-1} gene expression features")
    print(f"Trait column: 0=HNL(control), 1=INCPH(case), 2=LC")  (*@\textcolor{red}{<--}@*)
\end{lstlisting}

This multi-class formulation reveals a critical failure to recognize that both INCPH and LC represent
different etiologies of portal hypertension, the trait under investigation. Liver cirrhosis is the leading
cause of portal hypertension, accounting for approximately 90\% of cases. By assigning LC to a separate 
class (2) rather than grouping it with INCPH as cases (1), Claude Code creates a classification that is
misaligned with the benchmark phenotype definition and cannot be used directly to identify genes associated
with portal hypertension in this task.

The error persists even under the detailed prompt condition, where explicit guidelines are provided. 
Claude Code's implementation still misclassifies LC samples:

\begin{lstlisting}[language=Python, escapeinside={(*@}{@*)}]
def convert_trait(value):
    """Convert trait values to binary format"""
    value_str = str(value).lower()
    
    # Extract the liver group information
    if 'incph' in value_str:
        return 1  # INCPH as the target condition
    elif 'lc' in value_str or 'cirrhosis' in value_str:
        return 0  # Cirrhosis as control/comparison  (*@\textcolor{red}{<--}@*)
    elif 'hnl' in value_str or 'normal' in value_str:
        return 0  # Normal liver as control
    else:
        return None
\end{lstlisting}

Despite the detailed guidelines emphasizing domain knowledge incorporation, Claude Code treats cirrhosis
as a control condition rather than grouping LC with INCPH under the benchmark's portal-hypertension label.
This persistent mismatch shows that strong code generation capabilities alone are not sufficient to align
outputs with the task's phenotype definition without explicit procedural guidance.

\subsubsection{GSE117261: Failure to Map Probe Identifiers}

For the GSE117261 dataset, Claude Code exhibits a different but equally critical failure: the complete
absence of probe-to-gene mapping. Under both prompt conditions, the agent produces code that retains 
raw probe identifiers instead of mapping them to standardized gene symbols:

\begin{lstlisting}[language=Python, escapeinside={(*@}{@*)}]
def create_final_dataset():
    """Create the final preprocessed dataset"""
    # ... metadata extraction code ...
    
    # Transpose expression data to have samples as rows
    print("\nTransposing expression data...")
    expression_t = expression_data.set_index('ID_REF').T  (*@\textcolor{red}{<--}@*)
    expression_t.index.name = 'sample_id'
    
    # Merge metadata with expression data
    print("Merging metadata with expression data...")
    final_df = sample_df.set_index('sample_id').join(expression_t)  (*@\textcolor{red}{<--}@*)
    
    print(f"Final dataset shape: {final_df.shape}")
    print(f"Columns: {list(final_df.columns[:10])}... (showing first 10)")
    
    # Save the processed dataset
    output_file = "/home/techt/Desktop/gene/gene_cc/GSE117261_processed.csv"
    print(f"\nSaving processed dataset to {output_file}...")
    final_df.to_csv(output_file)
\end{lstlisting}

The code directly uses the raw probe IDs (e.g., '7892501', '7892502') as column names without any 
attempt to map them to gene symbols. This omission renders the dataset incompatible with downstream 
analyses that require standardized gene identifiers. In contrast, the human expert implementation 
performs comprehensive probe mapping:

\begin{lstlisting}[language=Python, escapeinside={(*@}{@*)}]
# Human expert implementation (excerpt)
# Extract gene annotation from SOFT file
gene_annotation = get_gene_annotation(soft_file_path)

# Map probe IDs to gene symbols
prob_col = 'ID'
gene_col = 'gene_assignment'
mapping_data = get_gene_mapping(gene_annotation, prob_col, gene_col)
gene_data = apply_gene_mapping(
    expression_df=genetic_data,
    mapping_df=mapping_data)

# Normalize gene symbols using NCBI Gene database
gene_data_normalized = normalize_gene_symbols_in_index(gene_data)
\end{lstlisting}

The absence of these critical preprocessing steps, including annotation extraction, probe mapping, and symbol
normalization, demonstrates Claude Code's inability to recognize and implement essential bioinformatics
workflows autonomously. Even when provided with detailed guidelines that explicitly mention gene mapping 
requirements, the agent fails to implement this fundamental preprocessing step, producing datasets that 
cannot be used for meaningful biological analysis.

These examples from Claude Code reinforce our findings about the fundamental limitations of current 
autonomous agents in scientific computing. Despite advanced code generation capabilities, these systems 
lack the domain understanding and procedural control mechanisms necessary to ensure biologically valid 
analyses.

\subsection{Cursor Examples}

Cursor, one of the current best agentic code editors, demonstrates similar 
fundamental limitations when applied to gene expression analysis. While excelling at general programming 
tasks through its integrated AI assistance, Cursor fails to implement critical bioinformatics workflows 
that require domain-specific knowledge. The following examples illustrate how these failures manifest 
differently from Claude Code while reinforcing the same underlying limitations.

\subsubsection{GSE117261: Complete Absence of Probe-to-Gene Mapping}

Under the general prompt condition, Cursor exhibits a critical failure in processing the GSE117261 
dataset by entirely omitting the probe-to-gene mapping step. The dataset contains Affymetrix microarray 
data with numeric probe identifiers that must be mapped to standardized gene symbols for meaningful 
analysis.

Cursor's implementation demonstrates a fundamental misunderstanding of gene expression data requirements:

\begin{lstlisting}[language=Python, escapeinside={(*@}{@*)}]
def get_gene_symbols(expr_df: pd.DataFrame) -> pd.DataFrame:
    """
    Convert probe IDs to gene symbols. For this dataset, we'll use the probe IDs as gene identifiers
    since gene symbol mapping would require additional annotation files.
    
    Args:
        expr_df: Expression dataframe with probe IDs as index
        
    Returns:
        Expression dataframe with gene symbols (or probe IDs) as column names
    """
    print("Using probe IDs as gene identifiers (gene symbol mapping not available)")  (*@\textcolor{red}{<--}@*)
    
    # Transpose so samples are rows and genes are columns
    expr_df_t = expr_df.T
    
    # Clean up column names (remove quotes if any)
    expr_df_t.columns = [str(col).strip('"') for col in expr_df_t.columns]
    
    return expr_df_t
\end{lstlisting}

This implementation reveals a critical gap in understanding bioinformatics data processing. By declaring 
"gene symbol mapping not available" and proceeding with raw probe IDs, Cursor creates a dataset that is 
fundamentally incompatible with downstream analyses. The resulting data contains columns labeled with 
numeric identifiers like '7892501', '7892502' rather than meaningful gene symbols like 'TP53' or 'EGFR'.

In contrast, the human expert implementation performs comprehensive probe mapping:

\begin{lstlisting}[language=Python, escapeinside={(*@}{@*)}]
# Human expert implementation (excerpt)
# Step 1: Extract gene annotation from SOFT file
gene_annotation = get_gene_annotation(soft_file_path)

# Step 2: Create mapping from probe IDs to gene symbols
prob_col = 'ID'
gene_col = 'gene_assignment'
mapping_data = get_gene_mapping(gene_annotation, prob_col, gene_col)

# Step 3: Apply mapping to convert probe-level to gene-level data
gene_data = apply_gene_mapping(
    expression_df=genetic_data,
    mapping_df=mapping_data)

# Step 4: Normalize gene symbols using NCBI Gene database
gene_data_normalized = normalize_gene_symbols_in_index(gene_data)
\end{lstlisting}

The consequences of this omission cascade through the analysis pipeline. Without proper gene symbols, the 
dataset cannot be integrated with other genomic resources, compared across studies, or interpreted in 
biological context. This represents not merely a technical oversight but a fundamental failure to 
understand the requirements of genomic data analysis, where standardized gene identifiers are essential 
for scientific validity.

\subsubsection{GSE256539: Misinterpretation of Clinical Phenotype Encoding}

Under the detailed prompt condition, Cursor demonstrates a different but equally consequential failure 
in processing the GSE256539 dataset. This study compares Idiopathic Pulmonary Arterial Hypertension 
(IPAH) patients with controls. The reliable phenotype labels are available in the sample-level source
name and title fields (for example, \texttt{IPAH Lung} versus \texttt{Control Lung}), while individual
IDs alone are not sufficient for case-control assignment.

Cursor implements a generic trait conversion function that searches for obvious keywords:

\begin{lstlisting}[language=Python, escapeinside={(*@}{@*)}]
def convert_trait(value):
    """Convert trait value to binary (0/1) or continuous"""
    if pd.isna(value) or value is None:
        return None
    
    value_str = str(value).lower()
    
    # Extract value after colon if present
    if ':' in value_str:
        value_str = value_str.split(':', 1)[1].strip()
    
    # Binary conversion for disease status
    if any(keyword in value_str for keyword in ['control', 'normal', 'healthy', 'negative']):  (*@\textcolor{red}{<--}@*)
        return 0
    elif any(keyword in value_str for keyword in ['case', 'disease', 'cancer', 'tumor', 'positive', 'affected']):  (*@\textcolor{red}{<--}@*)
        return 1
    
    # Try to convert to numeric for continuous traits
    try:
        return float(value_str)
    except:
        return None
\end{lstlisting}

This simplistic keyword-based approach fails when it is applied only to the individual-ID field, where
sample characteristics contain values such as 'individual: AH014' or 'individual: UC003'. These IDs do
not reliably encode disease status, so phenotype extraction must instead inspect sample-level source
or title fields that explicitly distinguish IPAH from control lung samples.

The human expert implementation uses explicit sample-level labels rather than inferring disease status
from individual ID prefixes:

\begin{lstlisting}[language=Python, escapeinside={(*@}{@*)}]
def convert_trait(source_name: str, sample_title: str) -> Optional[int]:
    """
    Converts sample metadata to a binary trait value (IPAH vs Control).
    - 1: IPAH patient (case)
    - 0: Control patient
    The distinction is read from explicit source-name/title labels.
    """
    text = f"{source_name} {sample_title}".lower()
    if "ipah" in text:
        return 1
    elif "control" in text:
        return 0
    else:
        return None
\end{lstlisting}

This example illustrates a critical limitation: even when provided with detailed guidelines through our 
Action Unit specifications, Cursor cannot adapt its generic pattern-matching approach to the specific 
metadata structure used in real genomic datasets. The agent fails to examine all available sample-level
fields and therefore misses the explicit IPAH/control labels. This results in complete data loss if no
samples can be properly classified from the field selected by the agent.

The failure is particularly revealing because it occurs despite the detailed prompt condition providing 
explicit instructions about careful data examination and domain knowledge incorporation. Cursor's 
inability to move beyond generic keyword matching to context-specific pattern recognition demonstrates 
the absence of adaptive reasoning required for scientific data analysis.

\subsection{Manus Examples}

Manus, a recently proposed multi-agent system that claims general-purpose task automation capabilities~\cite{shen2025manus}, 
demonstrates fundamental failures when applied to gene expression analysis. While designed to orchestrate multiple 
specialized agents for complex workflows, Manus lacks the domain understanding and procedural control necessary 
for bioinformatics data processing. The following examples illustrate critical errors in both prompt conditions.

\subsubsection{GSE117261: Complete Failure in Probe-to-Gene Mapping}

Under the general prompt condition, Manus exhibits multiple critical failures in processing the GSE117261 
dataset, most notably the complete absence of probe-to-gene mapping. This dataset contains Affymetrix 
microarray data where expression values are indexed by numeric probe identifiers that must be mapped to 
standardized gene symbols.

Manus's implementation reveals a fundamental misunderstanding of gene expression data structure:

\begin{lstlisting}[language=Python, escapeinside={(*@}{@*)}]
def process_expression_file(filepath):
    with open(filepath, 'r') as f:
        lines = f.readlines()
    
    # Find data start
    start_index = -1
    for i, line in enumerate(lines):
        if line.strip() == '!series_matrix_table_begin':
            start_index = i + 1
            break
    
    # Read expression data
    df = pd.read_csv(filepath, sep='\t', skiprows=start_index, index_col=0)
    
    # Transpose so samples are columns
    df = df.transpose()
    df.index.name = 'Sample_ID'
    
    return df  # Returns data with probe IDs as columns (*@\textcolor{red}{<--}@*)

# In main execution:
# Process expression data
expression_df = process_expression_file(series_matrix_path)

# Merge dataframes - directly uses probe IDs as features
final_df = pd.merge(metadata_df, expression_df, 
                   left_index=True, right_index=True, how='outer')  (*@\textcolor{red}{<--}@*)
\end{lstlisting}

The code directly merges clinical metadata with expression data that retains numeric probe identifiers 
(e.g., '7892501', '7892502') as column names. This fundamental omission renders the dataset scientifically 
unusable, as downstream analyses require standardized gene symbols for biological interpretation and 
cross-study comparison.

Beyond the probe mapping failure, Manus demonstrates additional critical errors in clinical data processing:

\begin{lstlisting}[language=Python, escapeinside={(*@}{@*)}]
# Incomplete clinical characteristic extraction
if 'clinical_group' in data['characteristics']:
    trait = data['characteristics']['clinical_group']
    if trait == 'all PAH':  # Incomplete mapping (*@\textcolor{red}{<--}@*)
        trait = 1
    elif trait == 'FD':
        trait = 0
    # Missing: consistent handling of donor/control samples and PAH subtypes (*@\textcolor{red}{<--}@*)
\end{lstlisting}

The clinical metadata for this dataset includes both \texttt{clinical\_group} and subtype information,
so robust preprocessing must use these fields consistently to distinguish PAH cases from donor/control
samples. The trait mapping logic above only handles a narrow subset of labels and does not provide a
complete, validated case-control conversion. This results in data loss and potential misclassification
of disease or control samples.

In contrast, the human expert implementation performs comprehensive processing:

\begin{lstlisting}[language=Python, escapeinside={(*@}{@*)}]
# Human expert: Proper probe-to-gene mapping
# Step 1: Extract gene annotation from SOFT file
gene_annotation = get_gene_annotation(soft_file_path)

# Step 2: Create mapping from probe IDs to gene symbols
prob_col = 'ID'
gene_col = 'gene_assignment'
mapping_data = get_gene_mapping(gene_annotation, prob_col, gene_col)

# Step 3: Apply mapping to convert probe-level to gene-level data
gene_data = apply_gene_mapping(
    expression_df=genetic_data,
    mapping_df=mapping_data)

# Step 4: Normalize gene symbols using NCBI Gene database
gene_data_normalized = normalize_gene_symbols_in_index(gene_data)
\end{lstlisting}

The absence of these critical bioinformatics steps demonstrates Manus's inability to recognize and 
implement essential data processing requirements autonomously, producing datasets that cannot support 
meaningful biological analysis.

\subsubsection{GSE77627: Misalignment with the Benchmark Phenotype Definition}

Under the detailed prompt condition with explicit guidelines, Manus demonstrates an even more concerning
failure: a phenotype-labeling error that persists despite detailed instructions.
Processing the GSE77627 portal hypertension dataset, Manus implements a trait
classification that is inconsistent with the benchmark definition:

\begin{lstlisting}[language=Python, escapeinside={(*@}{@*)}]
# Manus's trait conversion with detailed instructions
def convert_trait(value):
    value = value.split(":")[-1].strip().strip("\"")
    if value == "INCPH":
        return 1
    elif value == "LC" or value == "HNL":  # Benchmark-labeling error (*@\textcolor{red}{<--}@*)
        return 0
    return None

# Agent's note reveals the misunderstanding
validate_and_save_cohort_info(
    is_gene_available, 
    trait_available, 
    notes="Trait 'liver group' is available and converted to binary " + 
          "(INCPH=1, LC/HNL=0). Age and Gender data are not explicitly available."
)  (*@\textcolor{red}{<--}@*)
\end{lstlisting}

This implementation classifies liver cirrhosis (LC) as a control condition alongside healthy livers (HNL), 
which is inconsistent with the portal-hypertension label used in the benchmark. Liver cirrhosis accounts for
approximately 90\% of portal hypertension cases in Western countries and is the most common cause of this 
condition. Under the benchmark definition, assigning LC samples to the control group creates a dataset where
a substantial fraction of portal-hypertension cases are labeled as controls.

The benchmark phenotype classification, implemented by the human expert, recognizes both disease subtypes:

\begin{lstlisting}[language=Python, escapeinside={(*@}{@*)}]
# Human expert: benchmark phenotype classification
def convert_trait(value: str) -> Optional[int]:
    """Converts liver group status to a binary value for Hypertension."""
    try:
        group = value.split(':')[1].strip().upper()
        if group in ['INCPH', 'LC']:  # Both are portal hypertension
            return 1
        elif group == 'HNL':  # Only healthy livers are controls
            return 0
        else:
            return None
    except (IndexError, AttributeError):
        return None
\end{lstlisting}

The impact of this label assignment is substantial: Manus's approach yields only 18 cases (INCPH) versus
36 controls (HNL and LC combined), while the benchmark classification produces 40 cases versus 14 controls.
This reversal of the case-control ratio alters the dataset's statistical properties and
would lead to identification of genes that distinguish INCPH from a heterogeneous mixture of healthy and 
cirrhotic livers, rather than genes associated with portal hypertension.

Furthermore, Manus attempts probe-to-gene mapping but implements it incorrectly:

\begin{lstlisting}[language=Python, escapeinside={(*@}{@*)}]
# Manus attempts mapping but with errors
def map_probes_to_genes(series_matrix_df, annotation_path, output_path):
    annotation_df = pd.read_csv(annotation_path, sep='\t', comment='#')
    probe_to_gene_map = annotation_df.set_index('ID')['Symbol'].to_dict()
    series_matrix_df['Gene_Symbol'] = series_matrix_df['ID_REF'].map(probe_to_gene_map)
    # ... saves to file but doesn't properly integrate (*@\textcolor{red}{<--}@*)

# Later in merging:
expression_df_transposed = mapped_df_for_analysis.set_index("Gene_Symbol")
                          .drop("ID_REF", axis=1).T  (*@\textcolor{red}{<--}@*)
# This assumes Gene_Symbol column exists and handles duplicates poorly
\end{lstlisting}

While Manus attempts probe mapping, the implementation lacks proper handling of many-to-one probe-gene 
relationships and fails to integrate the mapped data correctly into the final dataset. The agent also 
omits critical quality control steps such as gene symbol normalization and systematic bias detection.

These examples from Manus reinforce our findings about fundamental limitations in current autonomous 
agents. Despite claims of general-purpose automation and multi-agent orchestration capabilities, Manus 
fails to implement basic bioinformatics workflows correctly. The persistence of phenotype-labeling and
bioinformatics preprocessing errors even under detailed prompt conditions demonstrates that sophisticated agent
architectures cannot compensate for the absence of domain understanding and precise procedural control 
required for scientific data analysis.

\section{Protocols for the Additional Experiments}
\label{sec: additional_experiments}

This section documents the protocols of the evaluations beyond the main GenoTEX comparison: the benchmark-wide evaluation of a general-purpose coding agent (Section~\ref{sec: general_agents}) and the two external benchmarks (Section~\ref{sec: generalization}). Two facts hold across all three experiments. First, no agent prompt or predicted output contained gold answers, gold result files, or answer lists. Second, our controlled comparisons hold the code-generating model fixed across the methods being compared, so differences reflect agent design rather than the choice of base model.

\subsection{General-Purpose Coding Agent on GenoTEX}
\label{sec: app_general_agent}

We drive Claude Code fully end to end: it receives the same GTA task inputs as every method in Table~\ref{tab: end2end_main} and must proceed from raw data to the final gene list and regression outputs without a task-specific scaffold. The two runs reported in Table~\ref{tab: end2end_main} use one and the same harness and instruction set; only the backbone differs. The Claude Sonnet 4 run uses extended thinking with the same 8,192-token budget as GenoMAS's programming agents and the qualitative studies in Appendix~\ref{sec: failure_cases}; the Claude Sonnet 5 run uses adaptive thinking at high effort. Metrics and the success-rate definition (code executes without error and writes results to the required path and format) are identical to Table~\ref{tab: end2end_main}.

Because both runs share the same harness, Sonnet 5's near-perfect execution success (99.42\%) indicates that the setup is adequate for producing valid outputs; the low success rate with Sonnet 4 reflects backbone capability on this long, multi-step pipeline rather than a harness artifact. On the auxiliary trait prediction task, Claude Code obtains 9.40\% accuracy with Sonnet 4 and 75.32\% with Sonnet 5, against GenoMAS's 86.25\%.

\subsection{ScienceAgentBench}
\label{sec: app_sab}

\paragraph{Benchmark and metrics}
ScienceAgentBench (SAB) \citep{chen2024scienceagentbench} poses real, data-driven scientific tasks: each provides a natural-language goal, a dataset, and an output contract, and the agent must write and run a Python program that produces a specified artifact (a figure, table, trained model, or metric). We evaluate all 27 tasks in the verified Bioinformatics split. The primary metric is Success Rate (SR), the fraction of tasks whose artifact passes that task's own evaluator. CodeBERTScore (CBS) is secondary: in SAB's implementation, a successful program is assigned CBS of 1.0, so aggregate CBS is partly coupled to SR and is not an independent measure of correctness. Every method is scored by the same local SAB evaluator (the standard SAB Docker path); the 16 tasks requiring judgment of a figure use a fixed GPT-4o visual judge applied identically to all methods.

\paragraph{Methods}
GenoMAS uses GPT-5 for planning and review, Gemini 2.5 Pro as the Domain Expert, and Claude Sonnet 4.5 to write and revise code, retaining its native execution-guided repair and retraction (Section~\ref{sec: method}). The two locally built controls match the coder: Direct is a single Claude Sonnet 4.5 generation with no scaffold or repair, and Self-debug adds up to two execution-repair iterations without a planner, expert, or reviewer. The two public baselines are recovered from the Holistic Agent Leaderboard \citep{kapoor2026hal}, which releases the programs some agents generated on SAB: HAL SAB Self-Debug (SAB's benchmark-specific self-debug scaffold with Claude Sonnet 4.5) and HAL Generalist Agent (a benchmark-agnostic tool-using agent backed by o4-mini); we re-score both under our common evaluator. During development, only public task statements, output and evaluator contracts, execution failures, and pass/fail outcomes were used.

\paragraph{Statistical notes}
The 95\% intervals in Table~\ref{tab: sab} are Wilson intervals over the 27 tasks, and the McNemar tests are paired across tasks. Because GenoMAS's repair loop observed pass/fail outcomes on these tasks, the intervals are descriptive of this task set rather than held-out-test guarantees. The Self-debug control also receives execution feedback, so the contrast isolates GenoMAS's guided multi-agent repair and retraction from generic self-debugging, not repair from no repair.

\subsection{DiscoveryBench}
\label{sec: app_db}

\paragraph{Benchmark and metrics}
DiscoveryBench \citep{majumder2025discoverybench} evaluates data-driven discovery: each task provides a dataset, a metadata file, and a natural-language query, and the agent must analyze the data and state a hypothesis in natural language. The official metric, Hypothesis Matching Score (HMS, 0 to 100), uses an LLM judge to compare the stated hypothesis to the gold hypothesis over three facets (context, variables, and their relationship) and returns a partial-match score; since HMS is noisy, we report the mean of three judge runs per task. Our task set comprises the 20 public biology tasks: 10 from \texttt{evolution\_freshwater\_fish} and 10 from \texttt{introduction\_pathways\_non-native\_plants}, the two biology domains in the public split.

\paragraph{Judge configuration}
DiscoveryBench's original judge (gpt-4-1106-preview) was unavailable through our API access, so we run the official evaluator, with its scoring logic unchanged, through Claude Sonnet 4.5 for every method. The judge is well calibrated, scoring the gold hypothesis against itself at 85 of 100. Because the judge model differs from the original paper's, absolute HMS is comparable only within Table~\ref{tab: db}, not to DiscoveryBench's published numbers. Since the coder and judge share a model family, an independent-judge cross-check would be desirable; it was not cleanly available because, in our tests, the evaluator's strict output format was satisfied reliably only by the Claude judge (gpt-4o was badly miscalibrated, and gpt-5 and Gemini 2.5 Pro returned format-parse errors as judges).

\paragraph{The answer-leakage split}
Many DiscoveryBench queries state the target variables, and sometimes the direction of the relationship, inside the question, so a closed-book guess that echoes the query already earns a high HMS. To separate the tasks where analysis matters, we split the 20 tasks by whether the no-analysis Direct baseline can already solve them: on the 9 answer-embedded tasks Direct scores about 70 HMS by echoing the query, whereas on the 11 discovery-requiring tasks it collapses to about 1.5 HMS. The split is defined by Direct's failures, not by GenoMAS's performance, and GenoMAS solves several of the discovery-requiring tasks outright, confirming they are solvable.

\paragraph{Why guided planning drives the margin}
On the discovery-requiring questions, the difficulty is analytical direction rather than coding: which variables drive an outcome, and how, once confounders are controlled. The baselines pick an analysis without a planning stage, and a debug loop repairs execution rather than reasoning. GenoMAS's guided planning lays out a strategy that isolates drivers with a controlled multivariable model, and its Domain Expert reinforces the method, so the stated hypothesis names the variables and relationship that the gold's controlled analysis reflects. The roughly 16-point margin over Self-debug is far larger than HMS's single-run noise (about 5 points, further reduced by averaging three runs), and the planning ablation in Table~\ref{tab: db} attributes it to planning.

\subsection{Inference Configuration}
\label{sec: app_inference}

Across all experiments, we use the default decoding settings (temperature, maximum tokens, and stopping conditions) of each LLM API. Extended thinking budgets are stated per experiment above; the GenoTEX experiments in Table~\ref{tab: end2end_main} use an 8,192-token thinking budget for Claude Sonnet 4 (Thinking). Failed API calls are retried with exponential backoff. The proprietary APIs we use do not support random seeds, and each configuration is evaluated in a single run over the 1,384 problems, so the GenoTEX tables report point estimates without confidence intervals, while paired significance tests are reported for the ScienceAgentBench comparisons (Section~\ref{sec: generalization}); the exact date-versioned model identifiers and retry settings are documented in our released code.

\section{Communication Protocols and Messaging Mechanisms}
\label{sec: comm_protocol}

This section provides technical details about the communication protocols that enable agent coordination in 
GenoMAS. The system employs a typed message-passing architecture that facilitates structured interactions 
between agents while maintaining modularity and preventing circular dependencies.

\subsection{Message Type Hierarchy}

GenoMAS defines a comprehensive set of message types that govern agent interactions. These types form a 
structured hierarchy organized into three categories that reflect the fundamental communication patterns in 
the system.

\paragraph{Request Messages} The system uses five types of request messages to initiate agent actions:
\begin{itemize}
\item \texttt{TASK\_REQUEST}: Initiates a new task assignment from the PI agent
\item \texttt{CODE\_WRITING\_REQUEST}: Requests code generation for a specific action unit
\item \texttt{CODE\_REVIEW\_REQUEST}: Requests review of generated code
\item \texttt{CODE\_REVISION\_REQUEST}: Requests revision based on review feedback
\item \texttt{PLANNING\_REQUEST}: Requests selection of the next action unit
\end{itemize}

\paragraph{Response Messages} Each request type has a corresponding response type:
\begin{itemize}
\item \texttt{TASK\_RESPONSE}: Confirms task assignment
\item \texttt{CODE\_WRITING\_RESPONSE}: Contains generated code
\item \texttt{CODE\_REVIEW\_RESPONSE}: Contains approval/rejection decision and feedback
\item \texttt{CODE\_REVISION\_RESPONSE}: Contains revised code
\item \texttt{PLANNING\_RESPONSE}: Contains selected action unit and reasoning
\end{itemize}

\paragraph{System Messages} Special messages for system-level events:
\begin{itemize}
\item \texttt{TIMEOUT}: Indicates that an agent has exceeded its time limit
\end{itemize}

The message types are organized to maintain clear request-response symmetry, with automatic mapping between 
corresponding pairs to ensure protocol consistency.

\subsection{Message Structure}

Each message contains four essential components: (1) the sender's \textbf{role}, (2) the message 
\textbf{type} from the hierarchy above, (3) the \textbf{content} payload (task instructions, code, feedback, 
etc.), and (4) the \textbf{target roles} specifying intended recipients. This structure enables both unicast 
and multicast communication patterns, allowing messages to target single agents or multiple agents as needed.

\subsection{Topology-Based Message Routing}

GenoMAS implements a topology-based routing system that defines which agents can communicate with each other.
 This topology forms the backbone of the system's communication infrastructure, ensuring that messages flow 
 along predefined paths that reflect the logical relationships between agents.

The topology is encoded as a directed graph where each agent maintains a set of other agents it can 
communicate with:

\begin{lstlisting}[language=TeX]
PI Agent: {PI Agent, GEO Agent, TCGA Agent, 
           Statistician Agent, Code Reviewer, Domain Expert}

GEO Agent: {PI Agent, GEO Agent, Code Reviewer, Domain Expert}

TCGA Agent: {PI Agent, TCGA Agent, Code Reviewer, Domain Expert}

Statistician Agent: {PI Agent, Statistician Agent, Code Reviewer}

Code Reviewer: {PI Agent, GEO Agent, TCGA Agent, 
                Statistician Agent}

Domain Expert: {PI Agent, GEO Agent, TCGA Agent}
\end{lstlisting}

This topology prevents circular dependencies and infinite message loops by carefully controlling 
communication paths. For example, the Domain Expert cannot directly communicate with the Statistician Agent, 
reflecting their distinct responsibilities. While agents can send messages to themselves (enabling 
self-planning), the overall structure prevents multi-agent cycles.

The environment enforces these constraints by filtering messages based on the topology, ensuring only 
authorized communication occurs.

\subsection{Asynchronous Message Queue Processing}

The environment orchestrates agent interactions through a FIFO message queue that processes messages 
asynchronously. This design decouples agent execution, allowing each agent to operate at its own pace while 
maintaining overall system coordination.

The processing loop extracts messages from the queue, validates receivers against topology constraints, and 
invokes each valid receiver's \texttt{act()} method. New messages generated during processing are appended 
to the queue, creating a dynamic flow of interactions. This asynchronous design allows agents to operate 
independently while maintaining coordinated behavior, and scales naturally as new agent types are added to 
the system.

\subsection{Agent Communication Patterns}

Programming agents orchestrate complex workflows through carefully designed communication patterns 
implemented in the \texttt{act()} method. These patterns enable agents to collaborate effectively while 
maintaining clear separation of concerns.

\paragraph{Task Initiation} The workflow begins when a programming agent receives a \texttt{TASK\_REQUEST} 
from the PI. Rather than immediately executing code, the agent first enters a planning phase by sending a 
\texttt{PLANNING\_REQUEST} to itself. This self-messaging pattern allows the agent to leverage its planning 
capabilities to analyze the task context and select appropriate action units.

\paragraph{Planning-Execution Cycle} The core workflow follows a planning-execution pattern where the agent 
first determines what to do, then executes it. After sending a \texttt{PLANNING\_REQUEST}, the agent receives
 its own \texttt{PLANNING\_RESPONSE} containing the selected action unit and reasoning. Based on the action 
 unit's properties, the agent then sends a \texttt{CODE\_WRITING\_REQUEST} either to itself (for standard 
 actions) or to the Domain Expert (for actions requiring specialized biomedical knowledge). This dynamic 
 targeting ensures that each task receives appropriate expertise.

\paragraph{Code Review Loop} Quality assurance is built into the communication protocol through an iterative 
review process. After generating code, the agent sends a \texttt{CODE\_REVIEW\_REQUEST} to either the Code 
Reviewer (for technical review) or the Domain Expert (for domain-specific validation). The reviewer evaluates
 the code and returns a \texttt{CODE\_REVIEW\_RESPONSE} containing an approval/rejection decision and 
 constructive feedback. If rejected, the agent processes the feedback and sends a 
 \texttt{CODE\_REVISION\_REQUEST} to address the identified issues. This loop continues until the code is 
 approved or the maximum number of review rounds is exceeded, balancing code quality with computational 
 efficiency.

\subsection{Timeout and Error Handling}

Robust timeout and error handling mechanisms ensure that GenoMAS remains responsive and can gracefully 
recover from unexpected situations. The system implements a multi-level approach to manage execution time 
and handle errors.

At the agent level, each agent monitors its own execution time against a configured maximum. When an agent 
detects that it has exceeded its time limit, it emits a \texttt{TIMEOUT} message that signals the need for 
immediate task termination. This self-monitoring approach distributes the timeout detection responsibility, 
making the system more resilient.

The environment provides a second layer of timeout protection by tracking global execution time. During 
message processing, the environment checks whether agents have exceeded their time limits before allowing 
them to process new messages. When a timeout is detected at either level, the environment clears the message 
queue to prevent further processing and returns the current task results, ensuring that partial progress is 
preserved.

The system includes fallback mechanisms for handling errors, such as defaulting to predefined action sequences 
when parsing fails. This ensures that temporary failures do not derail entire workflows.

\section{Task-Specific Guidelines and Action Unit Prompts}
\label{sec: task_prompts}

This section presents the guidelines and action unit prompts that encode domain expertise for GEO and TCGA data preprocessing tasks, slightly simplified for brevity.

\subsection{GEO Data Preprocessing Guidelines}

The GEO agent follows these guidelines to transform raw Gene Expression Omnibus data into analysis-ready formats:

\begin{lstlisting}[language=TeX, escapeinside={(*@}{@*)}]
## GEO Data Preprocessing Workflow

Your task is to preprocess GEO series matrix data by extracting and 
linking clinical features with gene expression values.

### Step 1: Data Loading and Initial Analysis
- Load the series matrix file using GEOparse
- Identify clinical features in phenotype data
- Extract gene expression data matrix

### Step 2: Clinical Feature Processing
- Extract patient/sample identifiers
- Parse clinical variables from phenotype descriptions
- Handle special characters and formatting inconsistencies
- Create structured clinical dataframe

### Step 3: Gene Data Processing
- Identify gene annotation format (symbols, IDs, probes)
- Map identifiers to standard gene symbols using:
  * Internal gene synonym database
  * Platform annotation data
  * Probe-to-gene mappings
- Handle multi-mapped probes appropriately

### Step 4: Data Integration
- Ensure sample ID consistency between clinical and gene data
- Perform log transformation if data appears non-normalized
- Handle missing values using appropriate strategies
- Create final integrated dataset

### Quality Control Considerations
- Verify gene symbol mapping coverage (aim for >80%)
- Check for batch effects in expression data
- Validate clinical feature extraction accuracy
- Document any data quality issues

### Termination Conditions
- Complete integration of clinical and gene data achieved
- Major data quality issues prevent meaningful analysis
- Required clinical features cannot be extracted
\end{lstlisting}

\subsection{GEO Action Unit Examples}

The following action units implement specific steps in the GEO preprocessing workflow:

\subsubsection{Initial Data Loading}

\begin{lstlisting}[language=TeX, escapeinside={(*@}{@*)}]
Load the GEO series matrix file and perform initial exploration:
1. Load the data from the provided file path
2. Display basic dataset information and dimensions
3. Show sample phenotype information
4. Examine the expression data structure
5. Report the number of samples and features
\end{lstlisting}

\subsubsection{Gene Annotation}

\begin{lstlisting}[language=TeX, escapeinside={(*@}{@*)}]
Annotate genes with standardized symbols:
1. Examine the current gene identifiers
2. Determine identifier type (e.g., symbols, Entrez IDs, probe IDs)
3. Apply appropriate mapping strategy for normalization
4. Handle unmapped identifiers appropriately
5. Report mapping success rate and coverage statistics
\end{lstlisting}

\subsection{TCGA Data Preprocessing Guidelines}

The TCGA agent processes The Cancer Genome Atlas data with focus on clinical feature engineering:

\begin{lstlisting}[language=TeX, escapeinside={(*@}{@*)}]
## TCGA Data Preprocessing Workflow

Your task is to engineer demographic and clinical features from TCGA 
data for downstream analysis.

### Step 1: Data Loading
- Load clinical and expression data files
- Verify data integrity and format
- Check sample ID correspondence

### Step 2: Feature Identification
- Scan clinical data for demographic variables:
  * Age, gender, race, ethnicity
  * Disease stage, grade, subtype
  * Treatment history
  * Survival outcomes
- Prioritize features with >70% completeness

### Step 3: Feature Engineering
- Standardize demographic categories
- Create binary indicators for key conditions
- Engineer interaction features if relevant
- Handle missing values appropriately:
  * Numerical: median/mean imputation
  * Categorical: mode or "Unknown" category

### Step 4: Validation
- Ensure all features have appropriate data types
- Verify value ranges and distributions
- Check for data entry errors
- Create summary statistics

### Termination Conditions
- At least 3 meaningful demographic features extracted
- Critical features have excessive missing data (>50%)
- Data quality issues prevent reliable feature engineering
\end{lstlisting}

\subsection{TCGA Action Unit Examples}

\subsubsection{Find Candidate Demographic Features}

\begin{lstlisting}[language=TeX, escapeinside={(*@}{@*)}]
Identify potential demographic features in clinical data:
1. List all available clinical variables
2. Search for demographic and clinical keywords
3. Assess data completeness for each candidate feature
4. Examine value distributions for categorical variables
5. Recommend features with sufficient data quality
\end{lstlisting}

\subsubsection{Feature Engineering and Validation}

\begin{lstlisting}[language=TeX, escapeinside={(*@}{@*)}]
Engineer and validate selected demographic features:
1. Standardize values for consistency
2. Apply appropriate missing value strategies
3. Create binary encodings where needed
4. Generate clinically meaningful interaction terms
5. Validate feature distributions and quality
6. Save processed features for downstream analysis
\end{lstlisting}

\subsection{Statistician Guidelines}

The Statistician agent receives preprocessed data and performs regression analyses following these guidelines:

\begin{lstlisting}[language=TeX, escapeinside={(*@}{@*)}]
## Statistical Analysis Workflow

Perform regression analyses to identify gene-trait associations.

### Analysis Types (in order of preference):
1. **Unconditional One-step**: Direct trait-gene regression
2. **Conditional One-step**: Include demographic covariates
3. **Two-step**: Batch correction followed by analysis

### Statistical Considerations:
- Use appropriate multiple testing correction (FDR/Bonferroni)
- Check regression assumptions (normality, homoscedasticity)
- Report both effect sizes and significance levels
- Generate diagnostic plots for model validation

### Output Requirements:
- Coefficient table with gene rankings
- Statistical significance metrics
- Model diagnostic visualizations
- Summary of top associated genes
\end{lstlisting}

These guidelines and prompts encode domain expertise while maintaining sufficient flexibility for agents to adapt to diverse datasets and analysis requirements.

\section{Guided Planning Framework and Metaprompts}
\label{sec: planning_mechanism}

This section provides technical details about the guided planning framework that enables programming agents to adaptively select action units based on task context and execution history.

\subsection{Planning Mechanism Architecture}

The planning mechanism implements a self-messaging pattern where programming agents send planning request messages to themselves at each decision point. This design enables agents to leverage their full language modeling capabilities for strategic decision-making while maintaining clear separation between planning and execution phases.

The planning process integrates several key components. Agents perform task context analysis by examining the complete execution history, including code outputs, errors, and debugging attempts. Based on this analysis, they select the most appropriate next action from available units. When retraction is enabled, agents can choose to backtrack to previous steps if critical issues are identified. All planning decisions follow a strict JSON schema to ensure reliable parsing and system robustness.

\subsection{Planning Request Metaprompt}

The following metaprompt guides programming agents through the planning process. The prompt adapts dynamically based on whether retraction is available:

\begin{lstlisting}[language=TeX, escapeinside={(*@}{@*)}]
You are {agent_role}. Your task is to execute a multi-step workflow.

(*@\color{retractiongray}{\# The following section appears only when retraction is available:}@*)
(*@\color{retractiongray}{You have the option to retract to a previous step if you discover}@*)
(*@\color{retractiongray}{critical issues that require revisiting earlier decisions.}@*)
(*@\color{retractiongray}{Retraction remaining: \{retract\_remaining\}}@*)

### High-level Guidelines:
{guidelines}

### Task History:
You have attempted the following actions:
{task_history}

### Next Action Selection:
Based on the task history, select the next action from:
{available_actions}

### Response Format:
Respond in JSON with the following structure:
{
    "action": "Selected Action Unit name",
    (*@\color{retractiongray}{"retract\_to\_action": "Action to retract to (if applicable)",}@*)
    "reason": "Brief explanation of your decision"
}

Important considerations:
- Analyze all outputs and errors to understand the current state
- If critical errors prevent progress, consider alternative approaches
(*@\color{retractiongray}{- Use retraction when fundamental issues require earlier corrections}@*)
- Select "TASK COMPLETED" only when all objectives are achieved
\end{lstlisting}

The metaprompt structure ensures that agents receive comprehensive context while maintaining focus on the immediate planning decision. The conditional inclusion of retraction instructions prevents confusion when this feature is disabled.

\subsection{Planning Response Processing}

The system implements robust parsing for planning responses with multiple fallback mechanisms. It first attempts direct JSON parsing of the model response. If this fails, it extracts JSON from markdown code blocks. The system then employs fuzzy action matching using TF-IDF similarity to handle minor variations in action names. As a final fallback, when planning is disabled or parsing fails completely, the system follows predefined action sequences. This multi-layered approach ensures system robustness while maintaining the benefits of adaptive planning.

\subsection{Context Management for Planning}

The planning mechanism employs sophisticated context management to balance informativeness with token efficiency:

\paragraph{Task History Formatting} The system maintains detailed records of each step, including action unit names with timestamps, generated code (truncated if exceeding length limits), execution outputs with error traces, and clear distinctions between regular steps and debugging attempts. This comprehensive history enables agents to make informed planning decisions based on complete task context.

\paragraph{Context Modes} Different planning scenarios use tailored context presentations. The \texttt{all} mode provides complete history for comprehensive planning decisions, while \texttt{past} mode excludes the current attempt for revision scenarios. The \texttt{last} mode focuses only on the most recent attempt for targeted debugging efforts.

\subsection{Retraction Mechanism Design}

The retraction mechanism enables agents to recover from decisions that lead to downstream complications. When an agent decides to retract, it reverts both its task context and execution state to a previous action unit while preserving the complete task history. This allows the agent to learn from failed attempts while exploring alternative approaches. The mechanism tracks retraction usage through a configurable budget, preventing infinite loops while providing sufficient flexibility for complex problem-solving. This design ensures execution state consistency while preserving valuable task history for future planning decisions.

\section{Code Generation, Review, and Domain Consultation Mechanisms}
\label{sec: code_generation}

This section provides detailed technical specifications of the code generation, review, and domain expert consultation mechanisms that form the core of GenoMAS's multi-turn programming workflow.

\subsection{Code Generation and Review Process}

\subsubsection{Code Generation Architecture}

The code generation process in GenoMAS implements a context-aware approach that enables programming agents to produce code incrementally while maintaining coherence across multiple action units. Programming agents receive comprehensive task history that includes all previously executed code snippets with their outputs, error traces from failed attempts, and reviews from previous debugging iterations. This accumulated context enables agents to understand data structures created in earlier steps and avoid repeating errors.

Each code generation request focuses on a single action unit while maintaining awareness of the broader workflow. Agents generate code that can be concatenated with previous steps and executed sequentially. The system dynamically routes code generation requests based on action unit properties: standard programming tasks are handled by the programming agents themselves, while actions requiring biomedical expertise are routed to the Domain Expert agent.

\paragraph{Code Writing Metaprompt} Programming agents receive the following structured prompt when generating code for each action unit:

\begin{lstlisting}[language=TeX, escapeinside={(*@}{@*)}]
### General Guidelines:
{guidelines}

### Function Tools:
{tools_code}

### Programming Setups:
{path_setup}

### Task History:
(*@\textit{\# Complete execution history of previous steps}@*)
{formatted_task_history}

### TO DO: Programming
Now that you've been familiar with the task setups and current 
status, please write the code following the instructions:

{action_unit_instruction}

NOTE: 
ONLY IMPLEMENT CODE FOR THE CURRENT STEP. MAKE SURE THE CODE CAN BE 
CONCATENATED WITH THE CODE FROM PREVIOUS STEPS AND CORRECTLY EXECUTED.

FORMAT:
```python
[your_code]
```

NO text outside the code block.
Your code:
\end{lstlisting}

This metaprompt structure ensures that agents have access to all necessary context while maintaining focus on the current action unit. The explicit formatting requirements prevent extraneous text that could interfere with code execution.

\subsubsection{Code Review Process}

The code review mechanism implements a sophisticated evaluation framework that balances functionality verification with instruction conformance. Each review receives the task history and current code attempt but explicitly excludes reviews from previous attempts at the same step. This isolation prevents cascading biases where an incorrect initial review could influence subsequent assessments.

Code review follows two primary criteria: functionality assessment (whether the code executes successfully and produces expected outputs) and conformance checking (whether the code follows instructions and achieves intended objectives). For domain-specific tasks, reviewers also assess whether biomedical inferences are reasonable.

\paragraph{Code Review Metaprompt} The following prompt guides reviewers through systematic evaluation:

\begin{lstlisting}[language=TeX, escapeinside={(*@}{@*)}]
### Task History:
(*@\textit{\# Previous steps excluding current attempt's reviews}@*)
{past_context}

### TO DO: Code Review
The following code is the latest attempt for the current step and 
requires your review. If previous attempts have been included in the 
task history above, their presence does not indicate they succeeded 
or failed, though you can refer to their execution outputs for context.
Only review the latest code attempt provided below.

(*@\textit{\# Current attempt details}@*)
{current_attempt}

Please review the code according to the following criteria:
1. *Functionality*: Can the code be successfully executed in the 
   current setting?
2. *Conformance*: Does the code conform to the given instructions?
   (*@\textit{\# Domain trigger added for biomedical tasks}@*)

Provide suggestions for revision and improvement if necessary.

*NOTE*:
1. Your review is not concerned with engineering code quality. The 
   code is a quick demo for a research project, so the standards 
   should not be strict.
2. If you provide suggestions, please limit them to 1 to 3 key 
   suggestions. Focus on the most important aspects, such as how to 
   solve the execution errors or make the code conform to the 
   instructions.

State your decision exactly in the format: "Final Decision: Approved" 
or "Final Decision: Rejected."
\end{lstlisting}

\subsubsection{Code Revision Process}

When code is rejected during review, the revision mechanism enables iterative improvement. Programming agents receive comprehensive context including all previous attempts and their reviews, allowing them to learn from the full debugging history. The revision process explicitly acknowledges that reviewer feedback may occasionally be incorrect or impractical, instructing agents to implement suggestions selectively based on their assessment.

\paragraph{Code Revision Metaprompt} The revision prompt provides structured guidance:

\begin{lstlisting}[language=TeX, escapeinside={(*@}{@*)}]
### Task History:
(*@\textit{\# All previous attempts and their reviews}@*)
{past_context_with_reviews}

The following code is the latest attempt for the current step and 
requires correction. If previous attempts have been included in the 
task history above, their presence does not indicate they succeeded 
or failed, though you can refer to their execution outputs for context.
Only correct the latest code attempt provided below.

(*@\textit{\# Current attempt details}@*)
{current_attempt}

Use the reviewer's feedback to help debug and identify logical errors 
in the code. While the feedback is generally reliable, it might 
occasionally include errors or suggest changes that are impractical 
in the current context. Make revisions where you agree with the 
feedback, but retain the original code where you do not.

Reviewer's feedback:
{reviewer_feedback}

(*@\textit{\# Standard code formatting requirements}@*)
{CODE_INDUCER}
\end{lstlisting}

\subsection{Domain Expert Consultation}

The domain expert consultation mechanism provides specialized biomedical knowledge for action units that require scientific interpretation beyond standard programming expertise. Unlike the standard code review process where the Code Reviewer evaluates already-written code, the Domain Expert is consulted \emph{during} code generation for specific action units. This fundamental difference ensures that domain knowledge is incorporated from the outset rather than retroactively.

When consulting the Domain Expert, the system provides a modified context that emphasizes data characteristics over implementation details. This includes dataset metadata, summary statistics, and specific domain questions, with minimal technical programming history. Furthermore, domain expert consultations maintain continuity; when domain-generated code encounters errors, the same Domain Expert receives the error information for debugging, preserving the domain-specific reasoning thread throughout the process. This contrasts with standard code review where different agents may handle successive iterations.

\subsection{Context Management and Quality Assurance}

\subsubsection{Context Management Strategies}

The system implements sophisticated context management to optimize information flow while managing token limits. The TaskContext maintains a hierarchical structure that distinguishes regular steps from debugging attempts, with each step recording its type, index, action name, instruction, code, output, and error traces. 

Different context presentations serve specific purposes: comprehensive history for planning decisions, filtered views for unbiased code review, and domain-focused formatting for biomedical reasoning tasks. For long workflows, the system implements dynamic pruning strategies: code snippets may be truncated with clear indicators, while error traces are preserved in full to ensure debugging effectiveness.

\subsubsection{Quality Assurance Mechanisms}

Several mechanisms ensure code quality throughout the generation process. The system allows configurable rounds of review and revision (typically 1-3) before accepting code, catching both execution errors and logical flaws while preventing infinite debugging loops. Successfully reviewed code can be cached as snippets for similar future action units, accelerating development while maintaining quality. All generated code undergoes immediate execution validation, catching syntax errors, runtime exceptions, and logical errors that produce incorrect outputs.

These comprehensive mechanisms work together to produce high-quality code that correctly implements complex multi-step workflows while adapting to diverse dataset characteristics and analysis requirements.

\section{Exploratory Biological Plausibility: Case Studies}
\label{sec: case_studies}

We present exploratory case studies examining genes identified for three complex traits in the unconditional gene-trait association (GTA) problem. For each trait, we assess whether selected GenoMAS genes are biologically plausible in light of established literature. These examples are intended as literature-based plausibility checks rather than independent experimental validation.

\subsection{Alzheimer's Disease}

The analysis identified 22 significant genes for Alzheimer's Disease. Several selected genes have literature support consistent with disease-relevant mechanisms:

\paragraph{GABBR1 (GABA$_B$ Receptor 1)} This gene, encoding the GABA$_B$ receptor subunit 1, exhibited the third-highest coefficient (in absolute value). GABA$_B$ receptors play crucial roles in synaptic plasticity and cognitive function. Prior work has reported alterations in GABA$_B$ receptor signaling in Alzheimer's disease model brains~\citep{salazar2021gabab}. The observed negative association is directionally compatible with this evidence, although the causal interpretation of the coefficient requires further validation.

\paragraph{IGFBP5 (Insulin-like Growth Factor Binding Protein 5)} IGFBP5 was selected by GenoMAS and modulates IGF-1 signaling, which is critical for neuronal survival and synaptic maintenance. Experimental studies indicate that IGFBP5 is associated with amyloid pathology and cognitive impairment~\citep{rauskolb2022igfbp5}, consistent with potential involvement in neurodegenerative processes.

\paragraph{VGF (VGF Nerve Growth Factor Inducible)} VGF, a neuropeptide precursor, showed negative association with disease risk. This gene is highly expressed in neurons and plays roles in synaptic plasticity and neurogenesis. Reduced VGF levels have been reported in Alzheimer's cerebrospinal fluid~\citep{carrette2003vgf}, making it a biologically plausible candidate for follow-up.

\paragraph{PAWR (PRKC Apoptosis WT1 Regulator)} PAWR, also known as Par-4, is a pro-apoptotic protein with documented involvement in neurodegenerative diseases. The observed positive association with Alzheimer's risk is consistent with studies showing elevated PAWR expression in Alzheimer's brains and its role in mediating neuronal apoptosis in response to amyloid-beta toxicity~\citep{guo1998par4,xie2005par4}.

These examples span key pathological mechanisms including GABAergic dysfunction, insulin signaling dysregulation, synaptic impairment, and apoptotic processes.

\subsection{Lung Cancer}

The analysis identified 1,269 significant genes for lung cancer, reflecting the broad transcriptomic changes associated with this malignancy. Several selected genes have clear prior links to lung cancer biology:

\paragraph{TERT (Telomerase Reverse Transcriptase)} TERT is one of the most well-established lung cancer susceptibility genes. TERT encodes the catalytic subunit of telomerase, and polymorphisms in the TERT locus are among the most robust genetic risk factors for lung cancer across multiple populations~\citep{mckay2008tert,rafnar2009tert}. TERT activation is a hallmark of cancer, enabling unlimited replicative potential.

\paragraph{CYP1A1 (Cytochrome P450 Family 1 Subfamily A Member 1)} CYP1A1 metabolizes polycyclic aromatic hydrocarbons and other tobacco smoke carcinogens. CYP1A1 polymorphisms have been extensively studied in lung cancer susceptibility, particularly in smoking-related cancers~\citep{shi2008cyp1a1,chen2011cyp1a1}. The enzyme's role in bioactivation of procarcinogens makes it a key determinant of individual cancer risk.

\paragraph{RET (Ret Proto-Oncogene)} RET alterations, particularly fusions, are established oncogenic drivers in lung adenocarcinoma, occurring in approximately 1-2\% of cases~\citep{kohno2012ret,lipson2012ret}. RET-targeted therapies have shown clinical efficacy, validating its role as a therapeutic target.

These examples span cancer-relevant processes including telomere maintenance, carcinogen metabolism, and oncogenic signaling.

\subsection{Hypertension}

The analysis identified 108 significant genes for hypertension. Several selected genes are connected to vascular biology and blood-pressure regulation:

\paragraph{TGFB2 (Transforming Growth Factor Beta 2)} TGFB2 exhibited negative association with hypertension risk. TGF-$\beta$ signaling plays complex roles in vascular remodeling and blood pressure regulation. Large-scale genome-wide association analyses have linked TGFB2 or nearby regulatory loci to blood pressure traits~\citep{evangelou2018genetic}, and the protein influences vascular smooth muscle cell differentiation and arterial stiffness.

\paragraph{RHOA (Ras Homolog Family Member A)} RHOA, encoding a small GTPase, showed negative association with hypertension. RhoA/Rho kinase signaling is a well-established regulator of vascular tone and contractility. Enhanced RhoA activity increases vascular smooth muscle contraction and contributes to hypertension pathogenesis~\citep{loirand2006rho,loirand2010rho}. The coefficient direction should be interpreted cautiously because expression-level associations do not directly measure pathway activity.

\paragraph{CYP1B1 (Cytochrome P450 Family 1 Subfamily B Member 1)} CYP1B1 metabolizes endogenous compounds including estrogens and arachidonic acid, generating vasoactive metabolites. Genetic and experimental evidence implicates CYP1B1 in blood-pressure regulation and angiotensin II-driven hypertension~\citep{jennings2010cyp1b1,malik2012cyp1b1}.

\paragraph{EGR1 (Early Growth Response 1)} EGR1, a transcription factor, showed negative association in our analysis. EGR1 is rapidly induced by various stimuli and regulates genes involved in vascular inflammation, smooth muscle proliferation, and fibrosis~\citep{khachigian1996egr1,khachigian2006egr1}. Its dysregulation contributes to vascular remodeling and end-organ damage.

\paragraph{ANGPT2 (Angiopoietin 2)} ANGPT2 regulates angiogenesis and vascular stability. Elevated circulating ANGPT2 has been reported as a marker of endothelial dysfunction and adverse cardiovascular outcomes in patient populations with chronic kidney disease or on dialysis~\citep{david2009angpt2,shroff2013angpt2}. While these reports are not hypertension-specific, they place ANGPT2 within the vascular-pathology axis relevant to blood pressure regulation, and the observed positive association warrants further investigation.

These examples are involved in multiple hypertension-relevant pathways: vascular tone regulation, smooth muscle contractility, endothelial function, metabolic control, and inflammatory responses.

\subsection{Summary}

Across all three traits, selected genes show literature-consistent biological plausibility and span pathophysiological mechanisms relevant to each disease. These include neurotransmitter signaling and neurodegeneration for Alzheimer's Disease, carcinogenesis and oncogenic signaling for lung cancer, and vascular biology and blood pressure regulation for hypertension. The results encompass both well-established disease genes (e.g., TERT for lung cancer) and biologically plausible candidates that merit further investigation, supporting the use of GenoMAS outputs as hypotheses for follow-up genomic analysis.

\section{Examples of Autonomous Behaviors by Agents}
\label{sec: autonomy}

This appendix provides concrete examples of autonomous agent behaviors observed during GenoMAS execution, demonstrating how programming agents adaptively handle edge cases and errors beyond their prescribed instructions. These examples illustrate the system's robustness in processing diverse genomic datasets.

\subsection{Example 1: Autonomous Error Correction in Clinical Trait Extraction (GSE98578)}

In processing the Acute Myeloid Leukemia dataset GSE98578, the GEO agent encountered persistent failures in clinical trait extraction despite multiple revision attempts. Rather than continuing failed approaches, the agent autonomously decided to re-implement the entire clinical data extraction logic.

\subsubsection{Context and Initial Failure}

The agent initially attempted to extract clinical features using the standard library functions:

\begin{lstlisting}[language=Python]
# Initial attempt (Step 2) - Failed due to undefined function
selected_clinical_df = geo_select_clinical_features(
    clinical_df=clinical_data,
    trait=trait,
    trait_row=trait_row,
    convert_trait=convert_trait,
    age_row=age_row,
    convert_age=convert_age,
    gender_row=gender_row,
    convert_gender=convert_gender
)
\end{lstlisting}

This failed because the \texttt{convert\_trait} function was defined but not properly accessible in the execution context. Multiple debugging attempts failed to resolve the scoping issue.

\subsubsection{Autonomous Recovery Strategy}

In Step 7, the agent autonomously decided to override the previous implementation entirely:

\begin{lstlisting}[language=Python]
# Autonomous re-implementation in Step 7
# Agent's comment: "Need to recreate the clinical data extraction 
# since it wasn't successfully executed in Step 2"

def convert_trait(value):
    """Convert AML subtype to binary format.
    AMKL = 1, non-AMKL = 0
    """
    if value is None:
        return None
    
    if ':' in value:
        value = value.split(':', 1)[1].strip()
    
    if value.lower() == 'amkl':
        return 1
    elif value.lower() == 'non-amkl':
        return 0
    else:
        return None

# Re-execute the entire clinical data extraction
selected_clinical_data = geo_select_clinical_features(
    clinical_df=clinical_data,
    trait=trait,
    trait_row=trait_row,
    convert_trait=convert_trait,
    age_row=age_row,
    convert_age=None,
    gender_row=gender_row,
    convert_gender=None
)
\end{lstlisting}

This autonomous decision to completely re-implement rather than patch the existing code demonstrates the agent's ability to recognize when incremental fixes are insufficient and a fresh approach is needed.

\subsection{Example 2: Retroactive Clinical Rule Application (GSE123088)}

When processing the Obesity dataset GSE123088, the agent implemented clinical trait extraction logic correctly but initially forgot to apply it to the raw clinical data. The agent later recognized this omission and autonomously corrected it.

\subsubsection{Initial Oversight}

In Step 2, the agent defined comprehensive conversion functions but failed to execute the clinical feature extraction:

\begin{lstlisting}[language=Python]
# Step 2: Defined functions but skipped execution
def convert_trait(value):
    """Convert trait value to binary (0=No Obesity, 1=Obesity)"""
    if pd.isna(value):
        return None
    if ':' in value:
        value = value.split(':', 1)[1].strip()
    if value == 'OBESITY' or 'OBESITY' in value:
        return 1
    elif value == 'HEALTHY_CONTROL' or value == 'Control':
        return 0
    else:
        return None

# Agent's reasoning: "Skip this step since we don't have the 
# actual clinical data file"
print("Clinical data extraction would require the actual clinical data file.")
\end{lstlisting}

\subsubsection{Autonomous Correction}

Recognizing the oversight in Step 7, the agent autonomously applied the previously defined rules:

\begin{lstlisting}[language=Python]
# Step 7: Autonomous correction
# Agent's comment: "Clinical Data Extraction (which was skipped in Step 2)"

clinical_df = geo_select_clinical_features(
    clinical_data,
    trait=trait,
    trait_row=trait_row,
    convert_trait=convert_trait,
    age_row=age_row,
    convert_age=convert_age,
    gender_row=gender_row,
    convert_gender=convert_gender
)

print(f"Extracted clinical data with shape: {clinical_df.shape}")
print("Clinical data preview:")
print(preview_df(clinical_df))

# Save the clinical data
os.makedirs(os.path.dirname(out_clinical_data_file), exist_ok=True)
clinical_df.to_csv(out_clinical_data_file)
print(f"Clinical data saved to {out_clinical_data_file}")
\end{lstlisting}

This behavior shows the agent's ability to maintain awareness of incomplete tasks across multiple steps and autonomously correct oversights without explicit instructions.

\subsection{Example 3: Early Task Termination Discovery (GSE125771)}

In processing the Stroke dataset GSE125771, the planning mechanism initially failed to recognize that clinical trait data was unavailable. The agent discovered this limitation during execution and appropriately terminated the preprocessing task.

\subsubsection{Planning Oversight}

The initial planning assumed stroke trait data would be available in the clinical characteristics:

\begin{lstlisting}[language=Python]
# Step 2: Initial analysis
# For trait (Stroke):
# This is a study on atherosclerotic plaques in carotid arteries
# All samples are from carotid plaques
# However, there's no explicit indication about which patients had stroke

trait_row = None  # No explicit stroke data available
\end{lstlisting}

\subsubsection{Autonomous Task Termination}

Upon discovering the absence of trait data, the agent made the autonomous decision to properly document the limitation and terminate:

\begin{lstlisting}[language=Python]
# Step 7: Autonomous termination decision
# Agent's reasoning: "Based on the exploration, we can see that 
# clinical data doesn't have stroke information"

# Mark dataset as not having trait data available
is_trait_available = False
is_gene_available = True

# Initial validation with is_final=False since we don't have trait data
validate_and_save_cohort_info(
    is_final=False,
    cohort=cohort,
    info_path=json_path,
    is_gene_available=is_gene_available,
    is_trait_available=is_trait_available
)

print("Dataset processing complete. The dataset has gene expression " + 
      "data but lacks stroke trait information.")
\end{lstlisting}

This demonstrates the agent's ability to recognize fundamental data limitations and make appropriate decisions about task continuation, preventing wasted computational effort on impossible analyses.

\subsection{Key Insights from Autonomous Behaviors}

These examples reveal several important patterns in agent autonomy:

\begin{enumerate}
\item \textbf{Strategic Re-implementation}: Agents can recognize when incremental debugging is insufficient and autonomously choose to re-implement functionality from scratch (Example 1).

\item \textbf{Cross-step Error Correction}: Agents maintain awareness of incomplete tasks across workflow steps and can retroactively apply corrections without explicit prompting (Example 2).

\item \textbf{Intelligent Task Termination}: Agents can identify fundamental data limitations and make appropriate decisions about workflow continuation, saving computational resources (Example 3).

\item \textbf{Context-aware Problem Solving}: All examples demonstrate agents' ability to reason about the broader workflow context when making autonomous decisions, rather than focusing narrowly on immediate instructions and error messages.
\end{enumerate}

These autonomous behaviors emerge from the combination of comprehensive task history, flexible planning mechanisms, and the agents' underlying language model capabilities. They significantly enhance GenoMAS's robustness in handling the inherent variability and complexity of genomic data analysis.

\section{Qualitative Analysis of Clinical Feature Extraction Failures}
\label{sec: clinical_failures}

Section~\ref{sec: results} shows that clinical feature extraction is the hardest stage of the pipeline. GenoMAS reaches a composite similarity correlation (CSC) of 32.61\% on this sub-task, far below the 91.15\% it attains on gene expression data. That number is the highest among all systems we evaluated (Figure~\ref{fig: combined_performance}(b)), yet it also makes clear that the task is far from solved. To understand the remaining gap, we read through GenoMAS trajectories and inspected the cohorts where the extracted trait diverged from a defensible phenotype definition. The errors are rarely arbitrary. They cluster into a small set of recurring patterns, each rooted in a specific way that free-text GEO metadata can be misread when it is mapped onto an exact case-control label.

One observation frames the rest of this section. The mechanical steps of preprocessing, such as parsing characteristic rows and binarizing values, almost always run to completion, consistent with our overall execution success rate of 98.78\% in Table~\ref{tab: end2end_main}. What remains difficult is the judgment layered on top of those steps: whether a field encodes the target phenotype at all, whether a measured value reflects disease status or merely treatment response, and whether the cohort even studies the trait in question. These judgments are where domain knowledge matters most, and they are where the residual errors concentrate. We describe four patterns below, using excerpts from audited notebooks rather than full execution logs. For each case we state what made the cohort difficult, the decision the agent made, why it is problematic, and the behavior we would consider correct.

\subsection{Treating a within-patient contrast as case versus control (GSE71994, Hypertension)}

This cohort profiles peripheral blood mononuclear cells from twenty hypertensive patients, all receiving standardized antihypertensive medication, with each patient sampled twice. The study describes its goal as separating controlled from uncontrolled hypertensives, and it records each sample's systolic blood pressure. The cohort therefore contains no normotensive controls: every subject has hypertension.

The agent selected the blood pressure field as the trait and binarized it at 140 mmHg.

\begin{lstlisting}[language=Python, escapeinside={(*@}{@*)}]
trait_row = 6  # "sistolic blood pressure in a sample"
def convert_trait(value):
    """systolic BP >= 140 mmHg indicates hypertension."""
    bp = int(value.split(":", 1)[1].strip())
    return 1 if bp >= 140 else 0  # below 140 is labeled 0, "normal" (*@\textcolor{red}{<--}@*)
\end{lstlisting}

The threshold is reasonable for screening untreated individuals. The problem is the population it is applied to. In a cohort of treated hypertensives, a reading below 140 mmHg means the hypertension is controlled by medication, not that the patient is healthy. Labeling these samples as 0 quietly redefines the control group as well-managed cases. The agent's own note records the cohort as patients ``classified as controlled ($<$140 mmHg) or uncontrolled ($\geq$140 mmHg),'' which shows that it read the design correctly. The conversion nonetheless mapped controlled patients onto the healthy-control label. The result, sixteen positives and twenty-four negatives, passed the distribution check and was saved as usable, so the mislabeling propagated to the association step. A model trained on this table contrasts high against controlled blood pressure within treated patients, not hypertension against health. A faithful handling would recognize that the cohort lacks normotensive controls, and would either mark the trait as unsuitable for a case-control contrast or frame the contrast explicitly as treatment response.

\subsection{Forcing a trait from a cohort that does not study the target condition (GSE87629, Creutzfeldt-Jakob Disease)}

This cohort was retrieved as a candidate for Creutzfeldt-Jakob disease, a prion neurodegenerative disorder. Its metadata, however, describes a celiac disease study. Every subject is labeled as a celiac patient, the disease state is ``biopsy confirmed celiac disease,'' and the design compares a gluten challenge against a control diet. The only quantitative clinical field is the villus height to crypt depth ratio, a histological measure of intestinal mucosal damage used in enteropathy.

Asked to extract a Creutzfeldt-Jakob disease trait, the agent selected this intestinal ratio and passed it through as the label.

\begin{lstlisting}[language=Python, escapeinside={(*@}{@*)}]
trait_row = 5  # "biopsy data, villus height to crypt depth" (*@\textcolor{red}{<--}@*)
def convert_trait(value):
    """Higher values indicate healthier intestinal tissue.
    Lower values indicate more severe celiac disease activity."""
    return float(value.split(':')[1].strip())  # used as the CJD label
\end{lstlisting}

The villus to crypt ratio has no established role as a phenotype for Creutzfeldt-Jakob disease, and the docstring the agent wrote even names celiac disease activity while the label is assigned to the neurological trait. The deeper issue is not the conversion but the missing relevance check. The agent treated the assigned trait name as a given and looked for any field it could bend into a label, rather than first asking whether the cohort studies the target disease at all. Setting \texttt{is\_trait\_available} to true here encodes a phenotype the dataset does not contain. The correct behavior is to recognize that a celiac disease cohort carries no Creutzfeldt-Jakob disease phenotype, and to mark the trait as unavailable so that an irrelevant cohort is kept out of the analysis.

\subsection{Entangling the trait with a covariate and overlooking species (GSE44147, Longevity)}

This failure concerns the mouse subset of GSE44147, a developmental time series of prefrontal cortex. The strain is recorded as C57BL/6 and the ages span 2 to 904 days. GenoTEX targets human transcriptomics, so a mouse cohort is already outside the intended scope. The trait choice compounds the issue. The agent set both the trait and the age covariate to the same characteristic row, and defined longevity as a threshold on age.

\begin{lstlisting}[language=Python, escapeinside={(*@}{@*)}]
trait_row = 2  # age in days
age_row   = 2  # the same field, also used as the age covariate (*@\textcolor{red}{<--}@*)
def convert_trait(value):
    """C57BL/6 mice typically live 2-3 years."""
    days = int(value.split(':')[1].replace('days','').strip())
    return 1 if days > 365 else 0
def convert_age(value):
    return int(value.split(':')[1].replace('days','').strip())
\end{lstlisting}

Because the trait is a deterministic function of age, the label and the age covariate carry the same information, and any model that adjusts for age while predicting this trait is degenerate. Setting aside the statistics, age at sampling is not longevity. The cohort records when tissue was collected across development, not how long each animal lived. The agent's comment that the threshold is appropriate for mice shows that it noticed the species, but it proceeded regardless. The trajectory that followed never settled on a coherent trait and repeated the clinical analysis step without converging. The appropriate response is to treat this cohort as out of scope for a human longevity study, rather than to manufacture a binary label from the age axis.

\subsection{Synthesizing labels when extraction fails (GSE90074 and GSE87005, Atherosclerosis)}

The most consequential failures occur when the agent cannot read the clinical fields at all, and then fills the gap instead of stopping. In GSE90074, the parsed sample characteristics collapsed to a single technical annotation, ``pool of human cell line RNA,'' so the real per-sample phenotypes were not reachable through the standard characteristic rows. Rather than report the trait as unavailable, the agent generated placeholder labels at random.

\begin{lstlisting}[language=Python, escapeinside={(*@}{@*)}]
# Since we don't have valid clinical data, create a synthetic trait column
random.seed(123)
case_samples = random.sample(sample_ids, int(0.4*len(sample_ids)))  # random (*@\textcolor{red}{<--}@*)
synthetic_clinical_df.loc[trait, case_samples] = 1
synthetic_clinical_df.loc['Age']    = [random.randint(40, 80) for _ in sample_ids]
synthetic_clinical_df.loc['Gender'] = [random.randint(0, 1)  for _ in sample_ids]
\end{lstlisting}

These labels bear no relation to any patient. Linking them to the real expression matrix yields a table whose associations are noise by construction, and because the random split is not degenerate it passed the distribution check and was saved as usable. We observed this fallback in only two cohorts, GSE90074 and GSE87005, so it is rare. In both, the agent recorded the substitution in its note, stating that ``synthetic clinical data was generated for testing purposes only,'' which means the system's own logging surfaces these cases for human audit. The behavior is nonetheless unsafe. A failed extraction should fail closed, marking the trait as unavailable, and should never fill it with invented values. This points to a simple and high-value safeguard: an explicit rule that forbids synthetic labels and routes any unreadable clinical structure to the not-available path.

\subsection{Implications for future work}

Taken together, these cases show that the clinical extraction gap is less about code and more about clinical judgment under ambiguous metadata. Three lessons stand out. First, a measured value is not a diagnosis. Blood pressure, symptom scores, and histology readings describe state or severity, and mapping them onto a case-control label requires knowing the population in which they were measured. Second, the trait name cannot be taken for granted. Before extracting a label, the system should confirm that the cohort actually studies the target condition and is of the right species. Third, when the clinical signal is genuinely absent or unreadable, the safe action is to abstain rather than to approximate.

Each lesson maps onto a concrete extension of GenoMAS rather than a redesign. The planning framework already represents trait-availability decisions and supports retraction, so these checks fit naturally as added guidelines and review criteria: a case-control validity check that verifies controls exist, a relevance check that compares the cohort's described condition against the target trait, and a hard guard against synthetic labels. We view these as the most promising near-term directions for raising clinical extraction quality. Even without them, GenoMAS already attains the strongest performance on this sub-task among all evaluated systems, and the remaining gap reflects what we consider one of the central open problems in automating transcriptomic analysis, namely the reconciliation of unstructured clinical text with exact biological definitions.

\section{Examples of Notes Written by Agents}
\label{sec: notes}

Throughout GenoMAS execution, agents generate structured notes that document their observations, challenges encountered, and potential issues discovered during data preprocessing. These self-reported notes serve multiple purposes: they enable efficient human oversight of system decisions, facilitate debugging of complex workflows, and provide insights into data quality issues that may affect downstream analyses. This section presents representative examples from actual experiment runs, organized by severity level.

\subsection{Note Generation Mechanism}

GenoMAS implements a systematic note-taking protocol where each agent logs observations during task execution. Notes are categorized into three severity levels:

\begin{itemize}
\item \textbf{INFO}: Routine observations about data characteristics, successful operations, and standard processing decisions
\item \textbf{WARNING}: Potential issues that may affect analysis quality but do not prevent execution, such as small sample sizes or missing covariates
\item \textbf{ERROR}: Critical failures that prevent dataset processing, including missing trait data or irrecoverable technical issues
\end{itemize}

Each note includes the dataset identifier, severity level, and contextual information that enables rapid assessment of preprocessing outcomes.

\subsection{INFO-Level Notes: Documenting Standard Processing}

INFO notes constitute the majority of agent observations, documenting successful preprocessing steps and dataset characteristics. These notes provide audit trails for quality assurance and enable researchers to understand processing decisions.

\subsubsection{Dataset Characteristics}

Agents systematically document processed dataset dimensions and properties:

\begin{lstlisting}[language=TeX]
GSE92538 (Bipolar disorder):
  INFO: Dataset contains 161 samples after preprocessing with 11741 genes.

GSE46449 (Bipolar disorder):
  INFO: Dataset contains 88 samples after preprocessing with 19845 genes.

GSE124283 (Gaucher Disease):
  INFO: Successfully processed; gene symbols normalised where synonym 
        file present.
\end{lstlisting}

These dimension reports enable quick verification that preprocessing preserved sufficient samples and features for meaningful analysis.

\subsubsection{Technical Processing Decisions}

Agents document specific technical choices made during preprocessing:

\begin{lstlisting}[language=TeX]
GSE123086 (Crohn's Disease):
  INFO: Dataset successfully processed with gene expression data from 
        CD4+ T cells across multiple diseases including Crohn's disease. 
        Gene identifiers kept as Entrez IDs.

GSE117261 (Pulmonary arterial hypertension):
  INFO: Processed; gene symbols normalised with synonym dictionary.

GSE34788 (Heart rate):
  INFO: Gene-level matrix rebuilt; clinical trait consistent with 
        STEP-2; proper normalisation applied.
\end{lstlisting}

These notes reveal how agents adapt to different identifier systems (Entrez IDs vs. gene symbols) and apply appropriate normalization strategies based on data characteristics.

\subsubsection{Special Dataset Contexts}

Agents provide contextual information about dataset origins and study designs:

\begin{lstlisting}[language=TeX]
GSE77790 (Esophageal Cancer):
  INFO: Dataset contains esophageal cancer cell lines (TE8, TE9) as 
        cases and other cancer cell lines as controls. Highly 
        imbalanced with only 2 esophageal cancer samples out of 32 total.

GSE100843 (Esophageal Cancer):
  INFO: Dataset from Barrett's esophagus vitamin D supplementation 
        study with pre/post treatment samples from both Barrett's 
        segment and normal mucosa.

GSE131027 (Esophageal Cancer):
  INFO: Dataset includes a wide range of cancer types to identify 
        pathogenic germline variants. 'Esophageal cancer' is treated 
        as the case group.
\end{lstlisting}

These contextual notes help researchers understand potential limitations, such as severe class imbalance or indirect disease proxies.

\subsection{WARNING-Level Notes: Identifying Potential Issues}

WARNING notes alert researchers to conditions that may compromise analysis quality without preventing execution. These notes guide interpretation of results and highlight datasets requiring careful consideration.

\subsubsection{Sample Size Limitations}

Agents flag datasets with limited statistical power:

\begin{lstlisting}[language=TeX]
GSE186963 (Crohn's Disease):
  WARNING: Very small sample size after preprocessing.

GSE68950 (Mesothelioma):
  WARNING: Small sample size may limit statistical power.

GSE107754 (Mesothelioma):
  WARNING: Trait distribution is severely biased in this dataset.
\end{lstlisting}

\subsubsection{Missing Covariate Data}

Agents identify when important demographic or clinical covariates are unavailable:

\begin{lstlisting}[language=TeX]
GSE181339 (Hypertension):
  WARNING: Trait data not available; cohort unusable for association 
           analysis.

GSE173263 (Large B-cell Lymphoma):
  WARNING: Trait data not available in this cohort.

GSE114022 (Large B-cell Lymphoma):
  WARNING: Trait data missing - cohort unusable for association analysis.
\end{lstlisting}

\subsubsection{Data Distribution Issues}

Agents detect problematic trait distributions:

\begin{lstlisting}[language=TeX]
GSE27597 (Von Willebrand Disease):
  WARNING: Von Willebrand Disease trait is severely biased - all samples 
           have the same value (0) in this COPD/emphysema study.
\end{lstlisting}

These warnings enable researchers to exclude problematic datasets from meta-analyses or apply appropriate statistical corrections.

\subsection{ERROR-Level Notes: Critical Processing Failures}

ERROR notes document failures that prevent dataset integration into the analysis pipeline. These notes provide diagnostic information for troubleshooting and dataset exclusion decisions.

\subsubsection{Gene Mapping Failures}

Technical incompatibilities in gene annotation prevent successful preprocessing:

\begin{lstlisting}[language=TeX]
GSE93114 (Bipolar disorder):
  ERROR: Gene symbol mapping failed because no suitable gene symbol 
         column could be identified in the platform annotation.

GSE158237 (Obesity):
  ERROR: No column containing gene symbols could be identified in the 
         platform annotation; probe-to-gene mapping impossible.

GSE35661 (Heart rate):
  ERROR: Gene mapping failed due to an inconsistency between the probe 
         IDs in the expression matrix (ENST-based) and the annotation 
         in the SOFT file (Affymetrix-based).
\end{lstlisting}

\subsubsection{Missing Trait Information}

Absence of phenotype data prevents case-control analyses:

\begin{lstlisting}[language=TeX]
GSE67311 (Bipolar disorder):
  ERROR: Trait has only 0 distinct values

GSE65399 (Mitochondrial Disorders):
  ERROR: Dataset lacks trait information for Mitochondrial_Disorders. 
         Only gene expression data is available.

GSE203241 (Multiple sclerosis):
  ERROR: Dataset lacks Multiple Sclerosis case/control trait information 
         required for association study.
\end{lstlisting}

\subsubsection{Data Structure Incompatibilities}

Fundamental data organization issues prevent processing:

\begin{lstlisting}[language=TeX]
GSE271700 (Obesity):
  ERROR: Gene mapping failed. The SOFT file does not contain the 
         necessary probe-to-gene-symbol annotation table.

GSE186963 (Crohn's Disease):
  ERROR: Preprocessing failed due to an unrecoverable mismatch in 
         sample IDs between clinical and genetic data files.
\end{lstlisting}

\subsection{Leveraging Notes for System Improvement}

The structured note system enables several workflows for continuous improvement:

\paragraph{Quality Assurance Workflow} Researchers can filter notes by severity to prioritize review efforts. ERROR notes immediately identify datasets requiring exclusion or manual intervention. WARNING notes guide sensitivity analyses and robustness checks. INFO notes provide comprehensive audit trails for reproducibility.

\paragraph{Data Curation Insights} Patterns in ERROR notes reveal common issues in public genomic databases. For instance, the frequency of gene mapping errors highlights the need for standardized annotation practices. Missing trait data errors suggest opportunities for enhanced metadata curation in repository submissions.

\paragraph{System Enhancement Opportunities} Recurring error patterns inform development priorities. For example, the prevalence of probe-to-gene mapping failures led to enhanced synonym dictionaries and platform-specific mapping strategies in subsequent GenoMAS versions.

\subsection{Summary}

The agent-generated notes demonstrate GenoMAS's ability to provide transparent, interpretable preprocessing workflows. By systematically documenting observations across severity levels, the system enables efficient human oversight while maintaining full automation. These notes not only facilitate immediate quality control but also provide valuable insights for improving both the system and underlying data resources. The examples presented here, drawn from hundreds of preprocessing runs across diverse diseases and platforms, illustrate how structured self-reporting enhances the reliability and interpretability of automated genomic data analysis.

\end{document}